\documentclass[letterpaper]{article} 
\usepackage{aaai2026}  
\usepackage{times}  
\usepackage{helvet}  
\usepackage{courier}  
\usepackage[hyphens]{url}  
\usepackage{graphicx} 
\usepackage{natbib}  
\usepackage{caption} 
\usepackage{booktabs}
\usepackage{amsmath}
\usepackage{enumitem}
\usepackage{subcaption}
\usepackage{amsfonts}
\usepackage{multirow}
\usepackage{adjustbox}
\DeclareMathOperator{\round}{round}
\usepackage{algorithm}
\usepackage{algorithmic}

\usepackage{newfloat}
\usepackage{listings}
\DeclareCaptionStyle{ruled}{labelfont=normalfont,labelsep=colon,strut=off} 
\floatstyle{ruled}
\newfloat{listing}{tb}{lst}{}
\floatname{listing}{Listing}
\title{Beyond Semantic Features: Pixel-level Mapping for Generalized AI-Generated Image Detection}

\author {
    Chenming Zhou\textsuperscript{\rm 1,\rm 2},
    Jiaan Wang\textsuperscript{\rm 1,\rm 2},
    Yu Li\textsuperscript{\rm 1,\rm 2}\thanks{Corresponding author.},
    Lei Li\textsuperscript{\rm 1,\rm 2},
    Juan Cao\textsuperscript{\rm 1,\rm 2},
    Sheng Tang\textsuperscript{\rm 1,\rm 2}
}

\affiliations {
    \textsuperscript{\rm 1}Institute of Computing Technology, Chinese Academy of Sciences, Beijing, China\\
    \textsuperscript{\rm 2}University of Chinese Academy of Sciences, Beijing, China\\
    \{zhouchenming21b,wangjiaan24s,liyu,lilei,caojuan,ts\}@ict.ac.cn
}

\begin{document}

\maketitle

\begin{abstract}
The rapid evolution of generative technologies necessitates reliable methods for detecting AI-generated images. A critical limitation of current detectors is their failure to generalize to images from unseen generative models, as they often overfit to source-specific semantic cues rather than learning universal generative artifacts. To overcome this, we introduce a simple yet remarkably effective \emph{pixel-level mapping} pre-processing step to disrupt the pixel value distribution of images and break the fragile, non-essential semantic patterns that detectors commonly exploit as shortcuts. This forces the detector to focus on more fundamental and generalizable high-frequency traces inherent to the image generation process. Through comprehensive experiments on GAN and diffusion-based generators, we show that our approach significantly boosts the cross-generator performance of state-of-the-art detectors. Extensive analysis further verifies our hypothesis that the disruption of semantic cues is the key to generalization.
\end{abstract}

\section{Introduction}
\label{intro}

The rapid advancement of generative models, from Generative Adversarial Networks (GANs) \cite{goodfellow2014generative} to contemporary Diffusion Models \cite{ho2020denoising}, has propelled AI-generated images with remarkable fidelity and diversity that are now virtually indistinguishable from authentic photographs. While these generative techniques have catalyzed innovation in creative arts and industrial applications, they have simultaneously precipitated a crisis of visual authenticity. The proliferation of AI-generated forgeries poses significant societal risks, particularly through the dissemination of misinformation and the erosion of evidentiary reliability in critical domains. This emerging threat landscape underscores the urgent need for robust detection methodologies in AI security.

In the task of in-distribution detection, classifiers often achieve remarkably high accuracy, yet their detection performance significantly deteriorates when faced with unknown generative models. To address the generalization challenge in detection, existing techniques typically fall into two categories: data-centric techniques that pre-process images to highlight forensic traces, and model-centric techniques that aim to learn more generalized features. \citet{wang2020cnn} demonstrate that aggressive data augmentation can enable convolutional classifiers to achieve cross-model generalization, while \citet{durall2020watch} analyze frequency-domain anomalies in GAN-generated images, identifying high-frequency artifacts caused by upsampling operations for generalized detection. Recent methods leverage the powerful feature learning capabilities of pre-trained large-scale models \cite{ojha2023towards, tan2025c2p, cozzolino2024raising, khan2024clipping}, assuming that their pre-training on vast real-image datasets allows them to detect synthetic content through deviations in feature distributions.

\begin{figure}[t]
  \centering
  \includegraphics[width=0.79\linewidth]{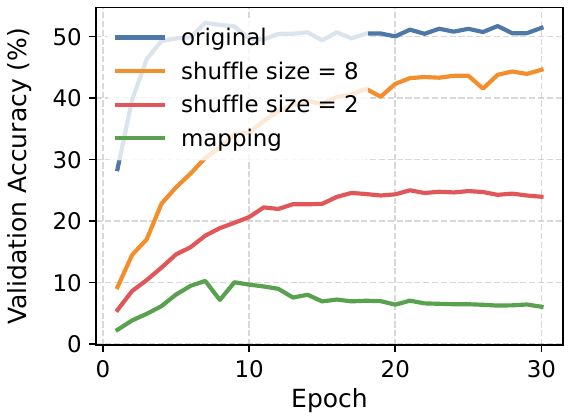}
  \caption{The impact of different image processing methods on ImageNet classification results.}
  \label{fig:imagenet}
\end{figure}

\begin{figure*}[t]
  \centering
  \includegraphics[width=1\linewidth]{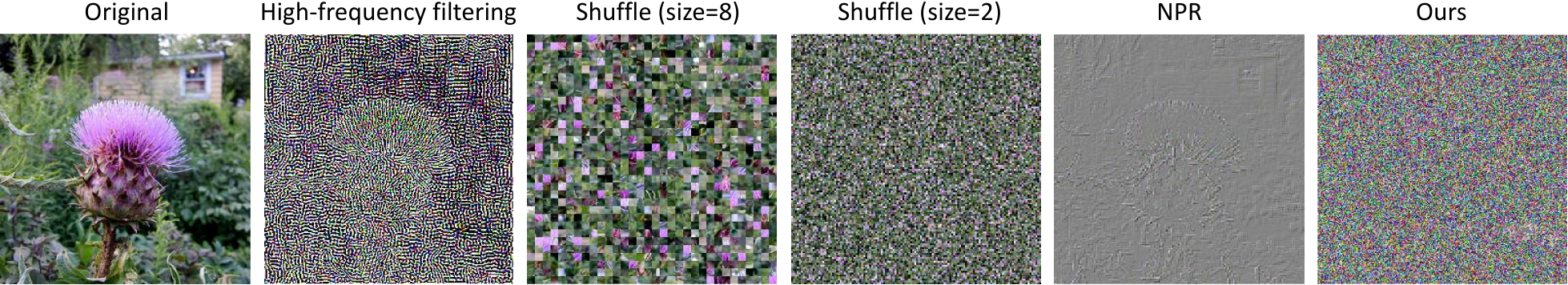}
  \caption{Visualization results of various semantic-reduction methods.}
  \label{fig:reduction}
\end{figure*}


Although existing methods generalize well when test and training distributions align, their accuracy declines significantly under substantial semantic shifts. This issue primarily stems from \emph{semantic bias} discrepancies caused by imperfect fitting to training data in generative models \cite{yanorthogonal, guillaro2025bias}, manifesting as visual artifacts like blurring and texture anomalies. However, different models exhibit specific semantic biases. As generative models continue to evolve with improved architectures and sampling techniques, these semantic biases diminish, leading to performance degradation in detectors that rely on them. Therefore, reducing the influence of semantic bias during classifier training is crucial for detection generalization.

To mitigate the impact of \emph{semantic bias}, some approaches leverage the Fourier transform to convert images into the frequency domain, where low-frequency components are removed via masking before reconstructing the images \cite{tan2024rethinking, bammey2023synthbuster, chu2024fire}. Although such spectral pruning can partially suppress low-frequency information, it fails to completely eliminate its interference and inevitably incurs information loss that degrades generative artifact detection. Alternative methods employ patch shuffling \cite{fu2025exploring, liu2022block, zheng2024breaking} to reduce the classifier's receptive field, thereby preventing overfitting to abstract semantic patterns. However, as shown in Figure \ref{fig:reduction}, high-pass filtering and NPR \cite{tan2024rethinking} methods can still preserve noticeable semantic information in images. Moreover, although the shuffle method progressively destroys semantics as the patch size decreases, the ImageNet classification experiments in Figure \ref{fig:imagenet} demonstrate that, despite slower convergence and lower validation accuracy, models still extract sufficient semantic information from shuffled patches even with the minimal patch size of 2.

To address the limitations of existing semantic-reduction methods, we propose a pixel-level mapping approach that reduces semantic bias through pixel value transformations. Since semantic bias mainly resides in low-frequency components, our method amplifies inter-pixel disparities to suppress low-frequency patterns while preserving detectable high-frequency artifacts. As shown in Figure \ref{fig:imagenet} and Figure \ref{fig:reduction}, our mapping achieves greater semantic reduction and lower validation accuracy than both baseline and shuffling methods, confirming the effective reduction of semantic information. By transforming low-frequency information, the impact of semantic bias on classifiers is significantly reduced, thereby amplifying the influence of high-frequency generative artifacts during training. While manipulating pixel values, our method preserves pixels' local correlations, offering two distinct advantages over existing bias mitigation strategies: 1) minimal information loss during image processing, and 2) enhanced emphasis on high-frequency forensic features during classifier training. Comprehensive experiments across diverse datasets and generative models (GANs and Diffusion models) demonstrate consistent improvements in cross-model generalization. Our key contributions are: 
\begin{itemize}[leftmargin=*]
    \item We empirically demonstrate that simple high-pass filtering and image patch shuffling fail to effectively eliminate semantic information, making them inadequate for reducing the impact of semantic bias on generalization in detection tasks. Our findings reveal critical limitations in existing approaches and highlight the need for more sophisticated semantic suppression methods.

    \item We introduce a novel pixel-level mapping method that attenuates semantic bias during classifier training. By transforming low-frequency information while amplifying high-frequency artifacts, our method significantly reduces classifier reliance on biased semantic features, addressing the core generalization challenge in synthetic image detection.

    \item Extensive validation across multiple benchmarks confirms the effectiveness of our method. The results show consistent performance gains when detecting images from unseen generative architectures, demonstrating the generalization capability of the proposed method.
\end{itemize}

\section{Related Work}

\subsection{Generative Models}
Generative models differ fundamentally from traditional autoencoders \cite{masci2011stacked, vincent2008extracting, salah2011contractive} by their ability to sample novel samples that conform to the distribution of existing data. GANs initially led synthetic image generation, with key variants addressing specific constraints: ProGAN \cite{karras2017progressive} for progressive training, StyleGAN \cite{karras2019style} for disentangled representations, BigGAN \cite{brock2018large} for large-scale synthesis, and StarGAN \cite{choi2018stargan} for multi-domain translation. Although GANs offer fast inference and modular extensibility, they produce noticeable semantic artifacts due to limited generation quality. Diffusion models \cite{ho2020denoising, song2020denoising, song2020score} later emerged as a mathematically grounded framework, overcoming GANs' training instability while ensuring better sample diversity. Recent large-scale implementations \cite{rombach2022high, ramesh2022hierarchical, saharia2022photorealistic} trained on extensive datasets achieve photorealistic high-resolution generation. These rapid advances in quality and diversity have markedly reduced semantic artifacts, thereby weakening detection methods dependent on semantic features.

\subsection{AI-Generated Image Detection}
The generalization challenge in synthetic image detection was first addressed by \citet{wang2020cnn}, demonstrating that aggressive data augmentation enables cross-GAN detection. \citet{durall2020watch} later revealed that GAN upsampling introduces distinct high-frequency artifacts, detectable via spectral analysis. However, such GAN-based observations limit classifier generalization. The emergence of diffusion models further challenged these approaches, as higher image fidelity eliminated many detectable artifacts. This prompted methods \cite{ojha2023towards, cozzolino2024raising, tan2025c2p} leveraging large-scale pretrained models to detect synthetic images by their deviations from natural distributions. Yet these models primarily rely on semantic features and overlook high-frequency traces correlated with generation artifacts. When test data diverges from fine-tuning distributions, their detection performance degrades significantly and shows inconsistency across domains.

To mitigate classifiers' over-reliance on semantic bias, existing methods disrupt global semantics via frequency manipulation or patch shuffling. Some studies \cite{tan2024rethinking, bammey2023synthbuster, chu2024fire} mitigate semantic interference by extracting high-frequency components through residual operations or frequency filtering. However, removing specific frequency bands cannot fully eliminate semantic influence, and the inherent coupling between generative artifacts and semantic content causes collateral damage under direct spectral suppression. While the patch shuffling strategies \cite{fu2025exploring, liu2022block, zheng2024breaking} aim to limit the classifier's receptive field by randomly permuting image patches, these semantic-suppression techniques force classifiers to prioritize forensic traces that generalize better across generator architectures. Yet, our experiments reveal that classifiers can still capture semantic information from shuffled patches through powerful fitting capacity. Meanwhile, the shuffling operation may disrupt the global structural patterns of generative artifacts.

\begin{figure*}[t]
  \centering
  \includegraphics[width=1\linewidth]{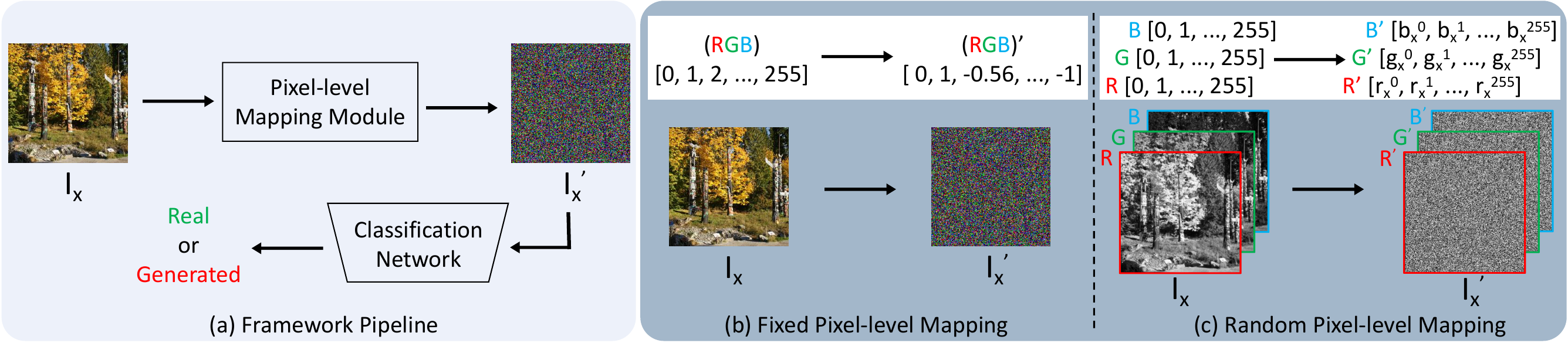}
  \caption{(a) The framework pipeline of the proposed method. The input image first passes through the pixel-level mapping module before being sent into the classification head. (b) The fixed pixel-level mapping module applies the same fixed mapping to all three channels of images. (c) The random pixel-level mapping module applies a different random mapping to the three channels of each image.}
  \label{fig:pipeline}
\end{figure*}

\section{Methodology}
\label{method}

\subsection{Preliminaries}

The detection generalization of generated images constitutes a classification task, aiming to perform binary classification on input images to determine whether they are synthesized by generative models \cite{wang2020cnn}. Specifically, the classifier is trained on images generated by a \textit{limited set} of known generative models, yet must maintain effectiveness when identifying images from \textit{unseen} generative architectures. Let $\mathcal{X} \subseteq \mathbb{R}^{H \times W \times C}$ denote the image space and $\mathcal{G} = \{G_1,...,G_k\}$ represent a set of known generative models. We define:
\begin{itemize}[leftmargin=*]
    \item \textbf{Training Data:} $\mathcal{D}_{\text{train}} = \{(x_i, y_i)\}_{i=1}^N$ where:
    \begin{equation}
        y_i = \begin{cases}
            1, & \text{if } x_i \sim P_{\text{gen}}(x|G_j), G_j \in \mathcal{G}\\
            0, & \text{if } x_i \sim P_{\text{real}}(x)
        \end{cases}
    \end{equation}
    
    \item \textbf{Objective:} Learn a classifier $f_\theta: \mathcal{X} \to [0,1]$ that minimizes the cross-entropy loss:
    \begin{equation}
        \mathcal{L} = -\frac{1}{N}\sum_{i=1}^N \left[y_i \log f_\theta(x_i) + (1-y_i) \log(1 - f_\theta(x_i))\right],
    \end{equation}
    while generalizing to unseen models $\mathcal{G}' = \{G_{k+1},...,G_{k+m}\}$.
\end{itemize}

Most existing studies on detection generalization focus on training data containing samples from only \textit{one category} of generative models \cite{wang2020cnn, frank2020leveraging, luo2024lare}. The trained classifier is required to:

\begin{enumerate}[leftmargin=*]
    \item Achieve \textit{intra-class generalization}: Maintain detection accuracy across:
    \begin{itemize}[leftmargin=*]
        \item Different architectures of the same model family, e.g., StyleGAN2 $\to$ BigGAN.
        \item Same model architecture trained with different data or hyperparameters, e.g., Stable Diffusion-V1.4 $\to$ Stable Diffusion-V1.5.
    \end{itemize}
    
    \item Demonstrate \textit{cross-class generalization}: Transfer effectively to:
    \begin{equation}
        \mathcal{G}_{\text{test}} \in \{\text{Diffusion Models}\} \times \{\text{GANs}\} \setminus \mathcal{G}_{\text{train}},
    \end{equation}
    where $\mathcal{G}_{\text{train}}$ contains only one model category.
\end{enumerate}

Training classifiers directly on RGB images can achieve high accuracy on generated images from the same source model, but performance significantly degrades on out-of-distribution data. This occurs because the classifier tends to overfit to semantic distribution patterns in the training set, limiting its generalization capability. Two primary factors contribute to this phenomenon: 1) Different generative models introduce distinct semantic artifacts (e.g., blurring, texture anomalies) due to variations in model architectures and training procedures - even identical models trained on the same data produce subtle differences from random initialization. 2) As generative models improve, their outputs increasingly approximate real data distributions, making semantic-level artifacts harder to detect in high-fidelity samples. To generalize better on unseen models, detectors must avoid relying on semantic bias in detection.

\subsection{Pixel-level Mapping for Semantic Bias Reduction}
\label{sec:mapping}

Generative artifacts in synthetic images exhibit tight coupling with semantic content, where perturbations to semantics inevitably distort artifact distributions. Existing works generally associate semantic bias with the low-frequency components of the smooth regions of the image, while generative traces are related to high-frequency details \cite{tan2024frequency, bammey2023synthbuster}. The inductive bias of convolutional classifiers towards low-frequency features \cite{tang2022defects} exacerbates semantic dominance during training. Therefore, weakening low-frequency semantic bias while enhancing high-frequency trace is a promising pathway to improve detection generalization. 

\begin{figure}[t]
  \centering
  \begin{subfigure}[b]{0.23\textwidth}
    \centering
    \includegraphics[width=\linewidth]{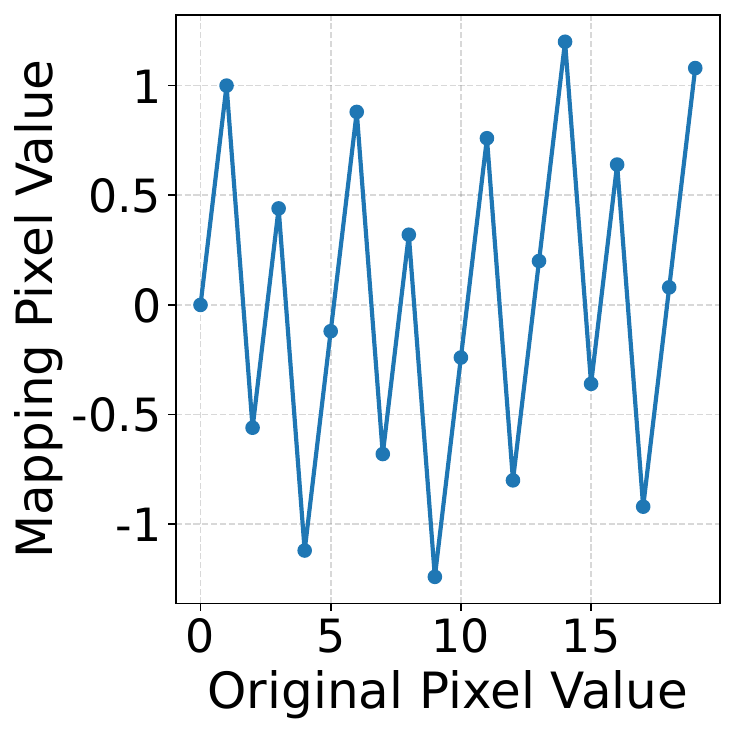}
    \caption{}
  \end{subfigure}
  \begin{subfigure}[b]{0.219\textwidth}
    \centering
    \includegraphics[width=\linewidth]{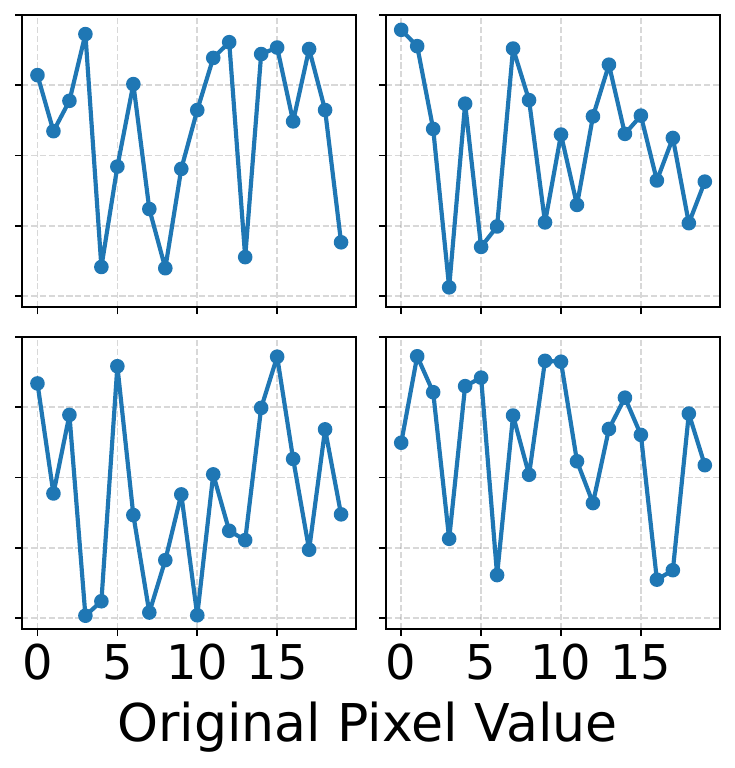}
    \caption{}
  \end{subfigure}
  \caption{(a) Fixed pixel-level mapping table, which remains the same for each channel of each sample. (b) Four examples of random pixel-level mapping tables, which maintain randomness for each channel of each sample.}
  \label{fig:mapping}
\end{figure}

To this end, we propose a pixel-level mapping approach that converts monotonically ordered pixel values (0-255) into a new set of pixel values, thereby altering the tight spatial arrangement between neighboring pixels. This transformation converts the image's low-frequency information into high-frequency while preserving the correlations between pixels. The overall process is illustrated in Figure \ref{fig:pipeline}(a). The input image undergoes semantic transformation through a pixel-level mapping module prior to being fed into the classification head. To effectively convert low-frequency information into high-frequency components by amplifying inter-pixel value differences, we propose a computationally efficient fixed mapping approach within our pixel-level mapping module in Figure \ref{fig:pipeline}(b). Our extended experimental results reveal an important insight: the specific pixel mapping relationship itself is not the critical factor. Rather, the key lies in disrupting the original monotonic pixel arrangement. Even when applying completely randomized mapping relationships that vary per sample as in Figure \ref{fig:pipeline}(c), the detection accuracy remains statistically comparable to deterministic mappings. Therefore, the fixed mapping proposed in this work can be viewed as one specific instantiation of this broader mapping paradigm. The implementation of the proposed pixel-level mapping module operates as follows:

\paragraph{Fixed pixel-level mapping module.} 
Given an input RGB image $I \in \mathbb{R}^{H \times W \times 3}$ with pixel values ranging from 0 to 255, we propose a mapping function that amplifies the differences between adjacent pixel values while normalizing the pixel intensities to facilitate classifier training. Formally, for each pixel value $v \in [0, 256)$, the mapping function $\phi_f$ can be expressed as:
\begin{equation}
\phi_f(v) = v - \round\left( \frac{v}{256}, 2 \right) \times 256,
\end{equation}
where $v$ denotes the original pixel intensity and $\round$ represents the round operation (equivalent to NumPy's \texttt{np.round} with \texttt{decimals=2}). The selection of \texttt{decimals} aligns with our monotonicity disruption principle. Setting \texttt{decimals=1} preserves pixel linearity due to coarse quantization (1/256$\approx$0.0039), while \texttt{decimals>1} effectively disrupt monotonic arrangements and \texttt{decimals=2} can simultaneously normalize the transformed pixel values to the approximate range of [-1.28, 1.28]. The correspondence between original and mapping pixel values in the fixed mapping module is illustrated in Figure \ref{fig:mapping}(a). To demonstrate the relation clearly, we present only the mapping results for pixel values in the range [0, 20]. As evident from the figure, the disparities between adjacent pixel values are significantly accentuated compared to the regularly normalized pixel values, thereby transforming originally smooth low-frequency regions in the image into high-frequency components.

\paragraph{Random pixel-level mapping module.} 
The fixed mapping method applies identical mapping rules to all input samples. Through further experiments, we observe that even when each sample (and every channel within each sample) undergoes transformations with randomly generated mapping tables, the classifier can still extract detection-generalizable features from the varying high-frequency information, achieving classification performance comparable to that of the fixed mapping approach. Concretely, for each input image $I \in \mathbb{R}^{H \times W \times 3}$, 
we construct a per-channel mapping table $T_c \in \mathbb{R}^{256}$ where each entry follows an independent and identically distributed (i.i.d.) uniform distribution over 256 dimensions as follows:
\begin{equation}
T_c \sim \mathcal{U}(-1,1)^{256}, \quad c \in \{0,1,2\}.
\end{equation}

The transformed image $I' \in \mathbb{R}^{H \times W \times 3}$ is then generated by:
\begin{equation}
I'_c[x, y] = T_c\big[I_c[x, y]\big], \quad \forall\ x \in [0,H), \ y \in [0,W),
\end{equation}
where $I_c[x, y] \in \{0,...,255\}$ denotes the original pixel value at location $(x, y)$ in channel $c$.

In Figure \ref{fig:mapping}(b), we visualize four randomly generated mapping tables. For clarity of presentation, only the mapped results for pixel values in the range of [0, 20] are displayed. The figure demonstrates that the randomized mappings can similarly amplify the disparities between adjacent pixels. Although distinct samples and image channels undergo different transformations, our subsequent experiments reveal that employing varied mappings still preserves the effectiveness of detection-relevant features.

\begin{table*}[t]
\centering
\begin{adjustbox}{width=\textwidth}
\begin{tabular}{lcccccccccc}
\toprule
Method & AttGAN & BEGAN & CramerGAN & InfoMaxGAN & MMDGAN & RelGAN & S3GAN & SNGAN & STGAN & \textbf{Mean} \\
\midrule
CNNDetection \cite{wang2020cnn} & 51.1 / 83.7 & 50.2 / 44.9 & 81.5 / 97.5 & 71.1 / 94.7 & 72.9 / 94.4 & 53.3 / 82.1 & 55.2 / 66.1 & 62.7 / 90.4 & 63.0 / 92.7 & 62.3 / 82.9\\
Frank \cite{frank2020leveraging} & 65.0 / 74.4 & 39.4 / 39.9 & 31.0 / 36.0 & 41.1 / 41.0 & 38.4 / 40.5 & 69.2 / 96.2 & 69.7 / 81.9 & 48.4 / 47.9 & 25.4 / 34.0 & 47.5 / 54.7\\
Durall\cite{durall2020watch} & 39.9 / 38.2 & 48.2 / 30.9 & 60.9 / 67.2 & 50.1 / 51.7 & 59.5/ 65.5 & 80.0 / 88.2 & \textbf{87.3} / \underline{97.0} & 54.8 / 58.9 & 62.1 / 72.5 & 60.3 / 63.3\\
Patchfor \cite{chai2020makes} & 68.0 / 92.9 & 97.1 / \textbf{100.0} & 97.8 / \underline{99.9} & 93.6 / 98.2 & 97.9 / \textbf{100.0} & 99.6 / \textbf{100.0} & 66.8 / 68.1 & 97.6 / \underline{99.8} & 92.7 / 99.8 & 90.1 / 95.4\\
F3Net \cite{qian2020thinking} & 85.2 / 94.8 & 87.1 / 97.5 & 89.5 / 99.8 & 67.1 / 83.1 & 73.7 / 99.6 & 98.8 / \textbf{100.0} & 65.4 / 70.0 & 51.6 / 93.6 & 60.3 / \underline{99.9} & 75.4 / 93.1\\
SelfBlend \cite{shiohara2022detecting} & 63.1 / 66.1 & 56.4 / 59.0 & 75.1 / 82.4 & 79.0 / 82.5 & 68.6 / 74.0 & 73.6 / 77.8 & 53.2 / 53.9 & 61.6 / 65.0 & 61.2 / 66.7 & 65.8 / 69.7\\
GANDetection \cite{mandelli2022detecting} & 57.4 / 75.1 & 67.9 / \textbf{100.0} & 67.8 / 99.7 & 67.6 / 92.4 & 67.7 / 99.3 & 60.9 / 86.2 & 69.6 / 83.5 & 66.7 / 90.6 & 69.6 / 97.2 & 66.1 / 91.6\\
LGrad \cite{tan2023learning} & 68.6 / 93.8 & 69.9 / 89.2 & 50.3 / 54.0 & 71.1 / 82.0 & 57.5 / 67.3 & 89.1 / \underline{99.1} & 78.5 / 86.0 & 78.0 / 87.4 & 54.8 / 68.0 & 68.6 / 80.8\\
UnivFD \cite{ojha2023towards} & 78.5 / 98.3 & 72.0 / 98.9 & 77.6 / 99.8 & 77.6 / 98.9 & 77.6 / 99.7 & 78.2 / 98.7 & \underline{85.2} / \textbf{98.1} & 77.6 / 98.7 & 74.2 / 97.8 & 77.6 / \textbf{98.8}\\
NPR \cite{tan2024rethinking} & 83.0 / 96.2 & 99.0 / \underline{99.8} & 98.7 / 99.0 & 94.5 / 98.3 & 98.6 / 99.0 & 99.6 / \textbf{100.0} & 79.0 / 80.0 & 88.8 / 97.4 & \underline{98.0} / \textbf{100.0} & 93.2 / 96.6\\
\midrule
Fixed-mapping & \textbf{99.6} / \underline{99.9} & \underline{99.1} / \textbf{100.0} & \underline{98.9} / \textbf{100.0} & \textbf{99.8} / \textbf{100.0} & \textbf{99.4} / \textbf{100.0} & \underline{99.7} / \textbf{100.0} & \underline{85.2} / 91.7 & \textbf{99.1} / \textbf{100.0} & \textbf{99.9} / \textbf{100.0} & \textbf{97.9} / \textbf{98.8}\\
Random-mapping & \underline{99.4} / \textbf{100.0} & \textbf{99.9} / \textbf{100.0} & \textbf{99.4} / 99.7 & \underline{99.0} / \underline{99.9} & \underline{99.2} / \underline{99.8} & \textbf{99.9} / \textbf{100.0} & 77.4 / 83.1 & \underline{98.3} / 99.8 & \textbf{99.9} / \textbf{100.0} & \underline{96.9} / \underline{98.0}\\
\bottomrule
\end{tabular}%
\end{adjustbox}
\caption{Cross-GAN performance (ACC./A.P.) analysis using the \textbf{Self-Synthesis} 9 GANs dataset, where \textbf{Bold} and \underline{underline} values denote the top and runner-up results.}
\label{tab:cross_gan_sources}
\end{table*}

\begin{table*}[t]
\centering
\begin{adjustbox}{width=\textwidth}
\begin{tabular}{lcccccccccc}
\toprule
Method & Ref & Midjourney & SDv1.4 & SDv1.5 & ADM & GLIDE & Wukong & VQDM & BigGAN & \textbf{mAcc} \\
\midrule
ResNet-50 \cite{he2016deep} & CVPR2016 & 54.9 & \underline{99.9} & 99.7 & 53.5 & 61.9 & 98.2 & 56.6 & 52.0 & 72.1 \\
DeiT-S \cite{touvron2021training} & ICML2021 & 55.6 & \underline{99.9} & \underline{99.8} & 49.8 & 58.1 & 98.9 & 56.9 & 53.5 & 71.6 \\
Swin-T \cite{liu2021swin} & ICCV2021 & 62.1 & \underline{99.9}& \underline{99.8} & 49.8 & 67.6 & 99.1 & 62.3 & 57.6 & 74.8 \\
CNNSpot \cite{wang2020cnn} & CVPR2020 & 52.8 & 96.3 & 95.9 & 50.1 & 39.8 & 78.6 & 53.4 & 46.8 & 64.2 \\
Spec \cite{zhang2019detecting} & WIFS2019 & 52.0 & 99.4 & 99.2 & 49.7 & 49.8 & 94.8 & 55.6 & 49.8 & 68.8 \\
F3Net \cite{qian2020thinking} & ECCV2020 & 50.1 & \underline{99.9} & \textbf{99.9} & 49.9 & 50.0 & \underline{99.9} & 49.9 & 49.9 & 68.7 \\
GramNet \cite{liu2020global} & CVPR2020 & 54.2 & 99.2 & 99.1 & 50.3 & 54.6 & 98.9 & 50.8 & 51.7 & 69.9 \\
UnivFD \cite{ojha2023towards} & CVPR2023 & 93.9 & 96.4 & 96.2 & 71.9 & 85.4 & 94.3 & 81.6 & 90.5 & 88.8 \\
DIRE \cite{wang2023dire} & ICCV2023 & 50.4 & \textbf{100.0} & \textbf{99.9} & 52.3 & 67.2 & \textbf{100.0} & 50.1 & 50.0 & 71.2 \\
FreqNet \cite{tan2024frequency} & AAAI2024 & 89.6 & 98.8 & 98.6 & 66.8 & 86.5 & 97.3 & 75.8 & 81.4 & 86.8 \\
NPR \cite{tan2024rethinking} & CVPR2024 & 81.0 & 98.2 & 97.9 & 76.9 & 89.8 & 96.9 & 84.1 & 84.2 & 88.6 \\
FatFormer \cite{liu2024forgery} & CVPR2024 & 92.7 & \textbf{100.0} & \textbf{99.9} & 75.9 & 88.0 & \underline{99.9} & \textbf{98.8} & 55.8 & 88.9 \\
DRCT \cite{chen2024drct} & ICML2024 & 91.5 & 95.0 & 94.4 & 79.4 & 89.2 & 94.7 & 90.0 & 81.7 & 89.5 \\
C2P-CLIP \cite{tan2025c2p} & AAAI2025 & 88.2 & 90.9 & 97.9 & 96.4 & \textbf{99.0} & 98.8 & 96.5 & \underline{98.7} & 95.8 \\
VIB-Net \cite{zhang2025towards} & CVPR2025 & 88.1 & 99.6 & 99.2 & 73.9 & 74.3 & 98.3 & 89.4 & 91.2 & 89.3 \\
B-Free \cite{guillaro2025bias} & CVPR2025 & 89.0 & 93.5 & 93.5 & 92.7 & 83.3 & 94.1 & 90.8 & 94.6 & 91.4 \\
Effort \cite{yan2024effort} & ICML2025 & 82.4 & 99.8 & \underline{99.8} & 78.7 & 93.3 & 97.4 & 91.7 & 77.6 & 91.1 \\
\midrule
Fixed-mapping & Ours& \textbf{96.8} & 98.9 & 98.8 & \textbf{98.7} & \underline{98.4} & 98.2 &\textbf{98.8} & \textbf{98.8} & \textbf{98.4} \\
Random-mapping & Ours & \underline{95.3} & 98.3& 97.6& \underline{98.0} & 97.2 & 97.5 &\underline{98.2} &98.0 & \underline{97.5}\\
\bottomrule
\end{tabular}%
\end{adjustbox}
\caption{Accuracy (Acc) comparison across models tested on \textbf{GenImage}, with training conducted on SDv1.4. \textbf{Bold} and \underline{underline} indicate top and runner-up performance respectively.}
\label{tab:cross_model_acc}
\end{table*}

\section{Experiments}
\label{exp}

\subsection{Setup}

\paragraph{Training datasets.}
We follow the settings from NPR \cite{tan2024rethinking} and C2P-CLIP \cite{tan2025c2p}, which use the ForenSynths \cite{wang2020cnn} and GenImage datasets \cite{zhu2023genimage}. The ForenSynths dataset contains 20 semantic classes, of which we exclusively employ 4 classes (i.e., car, cat, chair, horse) during training to maintain consistency with previous works. For the GenImage dataset, we adopt SDv1.4 \cite{rombach2022high} as the generative model. The real images in both ForenSynths and GenImage originate from the LSUN \cite{yu2015lsun} and ImageNet datasets \cite{russakovsky2015imagenet}.

\paragraph{Test datasets.}
To evaluate the generalization capability of the proposed method in real-world scenarios, we incorporate diverse real images alongside GAN and Diffusion models. The evaluation benchmark comprises two datasets of Self-Synthesis \cite{tan2024frequency} and GenImage \cite{zhu2023genimage}. The detailed descriptions are available in the Supplementary Materials.

\paragraph{Implementation details.}
\label{exp:imp}
Our method is implemented using PyTorch \cite{paszke1912imperative} with 8 Nvidia 3090 GPUs. We implement the detector network with a ResNet-50 \cite{he2016deep} architecture. During training, we randomly crop images to size 128$\times$128 to avoid resizing bias. During testing, we center-crop the images. We utilize the Adam optimizer \cite{kingma2014adam} with an initial learning rate of $2 \times 10^{-4}$. The first-order moment decay rate and the second-order moment decay rate are set to 0.9 and 0.999, respectively, and weight decay is set to $2 \times 10^{-4}$. We train the detector network for 200 epochs with a batch size of 128.

\paragraph{Metrics.}
We follow existing works \cite{ojha2023towards,liu2024forgery,tan2025c2p}, to compare and report classification accuracy (Acc) and average precision (AP). A uniform classification threshold of 0.5 is maintained on all evaluation benchmarks to ensure an equitable comparison of detection performance.

\subsection{Quantitative Analysis}
We present a comprehensive evaluation of our proposed method against competing approaches through cross-dataset and cross-model testing.

\paragraph{Evaluation on Self-Synthesis GAN dataset.}
The results of accuracy are shown in Table \ref{tab:cross_gan_sources}. The results of comparative methods are from NPR \cite{tan2024rethinking}. The training setting is the same as NPR which uses ProGAN (4 classes). This benchmark contains data from 9 state-of-the-art GAN models to evaluate generalization ability across different GAN models. Our method surpasses the baseline UniFD 
 \cite{ojha2023towards} by 20.3\% in classification accuracy and outperforms the current state-of-the-art NPR \cite{tan2024rethinking} by 4.7\%, demonstrating remarkable generalization over different GAN models.

\begin{table}[t]
\centering
\begin{tabular}{lcc}
\toprule
Method &  mAcc & mAP\\
\midrule
ResNet-50 & 67.0 & 76.9 \\
High-frequency filtering & 64.4 & 70.2 \\
Patch shuffling (size=8) & 70.7 & 80.6 \\
Patch shuffling (size=2) & 50.5 & 51.0 \\
BSA \cite{zheng2024breaking} & 74.7 & 85.6 \\
NPR \cite{tan2024rethinking} & 88.6 & 93.7 \\
\midrule
Fixed-mapping & \textbf{98.4} & \textbf{99.8} \\
Random-mapping & \underline{97.5} & \underline{99.6}\\
\bottomrule
\end{tabular}%
\caption{Comparison with different semantic reduction methods tested on \textbf{GenImage}, with training conducted on SDv1.4. \textbf{Bold} and \underline{underline} indicate top and runner-up performance respectively.}
\label{tab:semantic_reduction_comparison}
\end{table}

\paragraph{Evaluation on GenImage dataset.}
The results of accuracy are shown in Table \ref{tab:cross_model_acc}. The results of the compared methods are from GenImage \cite{zhu2023genimage}, C2P-CLIP \cite{tan2025c2p} and DRCT \cite{chen2024drct}. All the detection models are trained on the GenImage training set with SDv1.4 as the generative model. The GenImage dataset incorporates synthetic images generated by state-of-the-art diffusion models, including commercial closed-source models such as MidJourney and WuKong. Notably, a subset of generated images employs substantially higher resolutions. For instance, MidJourney outputs 1024×1024 pixel images, whose resolution discrepancy from conventional datasets introduces resolution bias that poses significant challenges to detection robustness. Our pixel-level mapping method achieves a new state-of-the-art result with an average accuracy of 98.4\%, exceeding baseline UniFD and state-of-the-art C2P-CLIP by 9.6\% and 2.6\% respectively. We additionally conduct experiments on the UniversalFakeDetect dataset \cite{ojha2023towards}, with detailed results provided in the Supplementary Materials.

\paragraph{Comparison with different semantic reduction methods.}
To evaluate the impact of different semantic reduction methods on detection performance, we conduct comparative experiments following the protocol in Table \ref{tab:cross_model_acc}. Our analysis includes: 1) baseline approaches (high-pass filtering and patch shuffling with sizes 8×8 and 2×2) implemented on ResNet-50, and 2) state-of-the-art variants - NPR (spectrum-based) \cite{tan2024rethinking} and BSA (shuffle-based) \cite{zheng2024breaking} - representing advanced improvements over these core ideas. For NPR, we directly report results from the original paper, while for BSA (which lacked GenImage benchmarks in its publication), we faithfully reproduced the method using the author's official codebase and report new evaluation results in Table \ref{tab:semantic_reduction_comparison}. The results in Table \ref{tab:semantic_reduction_comparison} reveal several key insights: 1) High-pass filtering underperforms the baseline due to excessive information loss from discarding low-frequency components, while residual semantic biases persist in mid-high frequencies; 2) While patch shuffling with size 8×8 demonstrates marginal performance gains, the extreme case of 2×2 patches proves counterproductive - the excessive fragmentation prevents meaningful feature learning, as evidenced by the classifier's failure in detection. 3) Both BSA and NPR demonstrate stronger performance through their specialized designs (receptive field restriction and residual operations respectively); 4) Our proposed pixel mapping approach achieves significant performance gains, validating its dual advantage in effectively suppressing semantic bias while preserving discriminative generative artifacts. Notably, the random mapping variant exhibits slightly inferior performance compared to fixed mapping, suggesting that consistent transformation patterns benefit training convergence despite equivalent semantic suppression.

\subsection{Qualitative Analysis}
We present visual comparisons of different semantic reduction approaches.

\begin{figure}[t]
  \centering
  \begin{subfigure}[t]{0.11\textwidth}
    \centering
  \includegraphics[width=\linewidth]{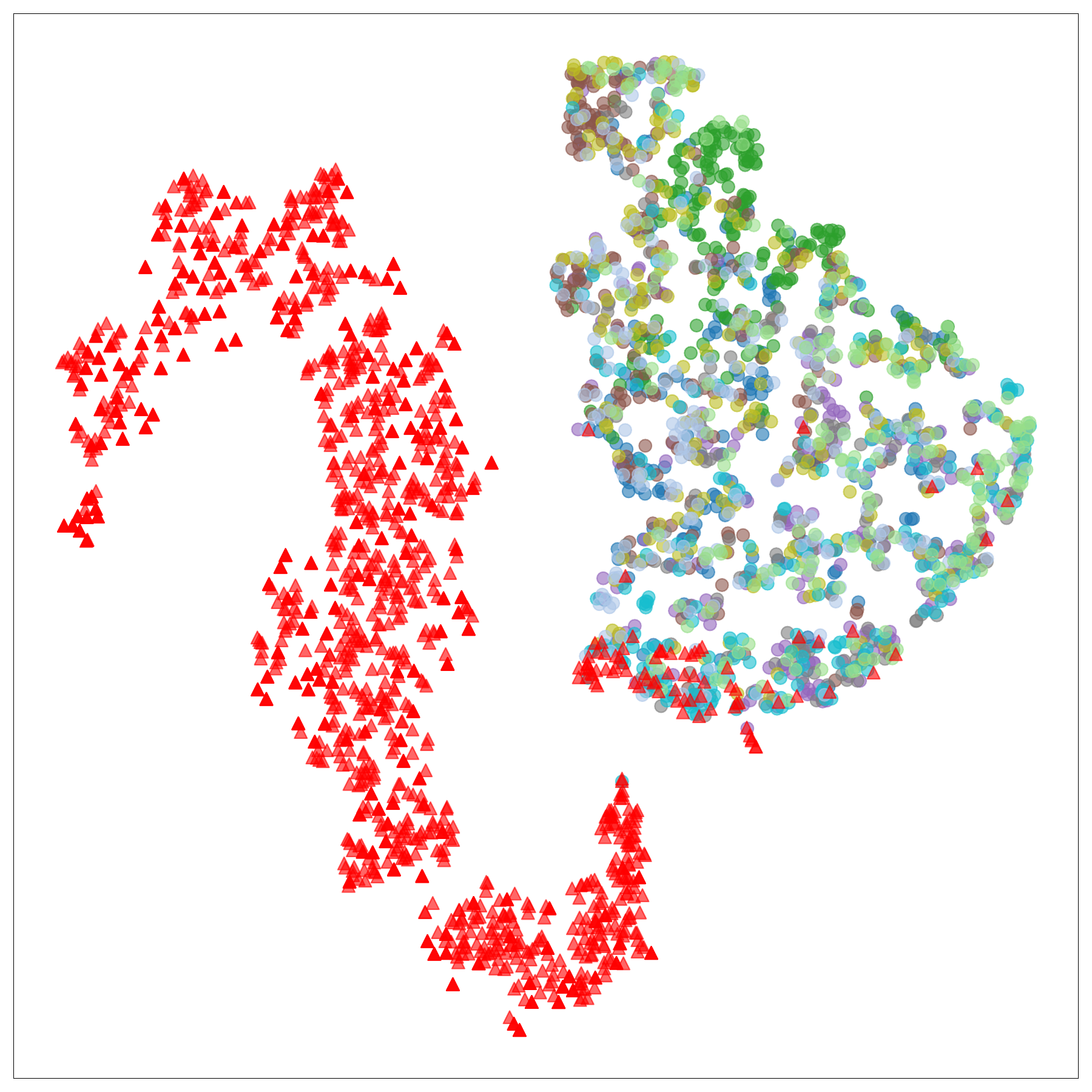}
  \caption{}
  \end{subfigure}
  \hfill
  \begin{subfigure}[t]{0.11\textwidth}
    \centering
  \includegraphics[width=\linewidth]{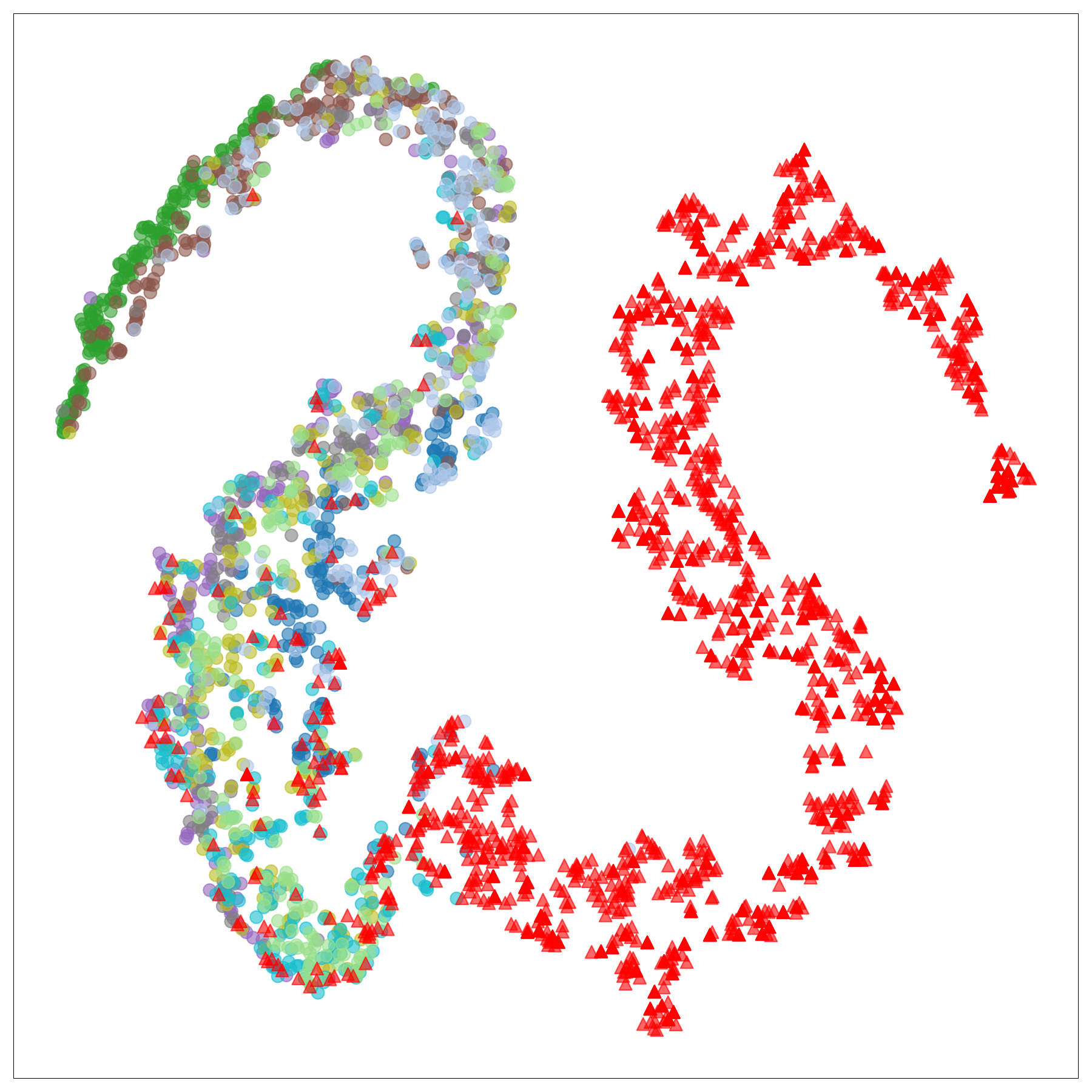}
  \caption{}
  \end{subfigure}
  \hfill
  \begin{subfigure}[t]{0.11\textwidth}
    \centering
  \includegraphics[width=\linewidth]{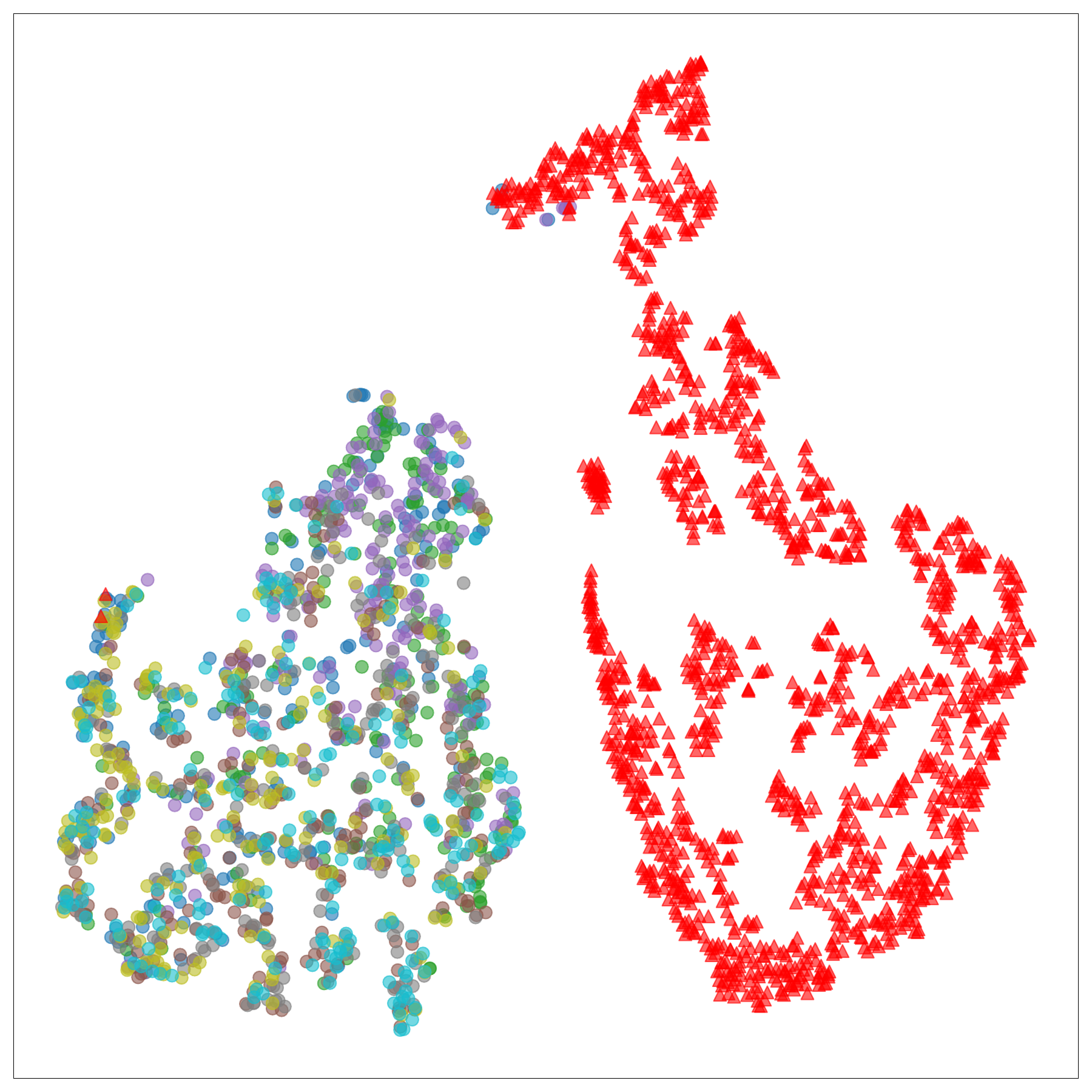}
  \caption{}
  \end{subfigure}
  \hfill
   \begin{subfigure}[t]{0.11\textwidth}
    \centering
  \includegraphics[width=\linewidth]{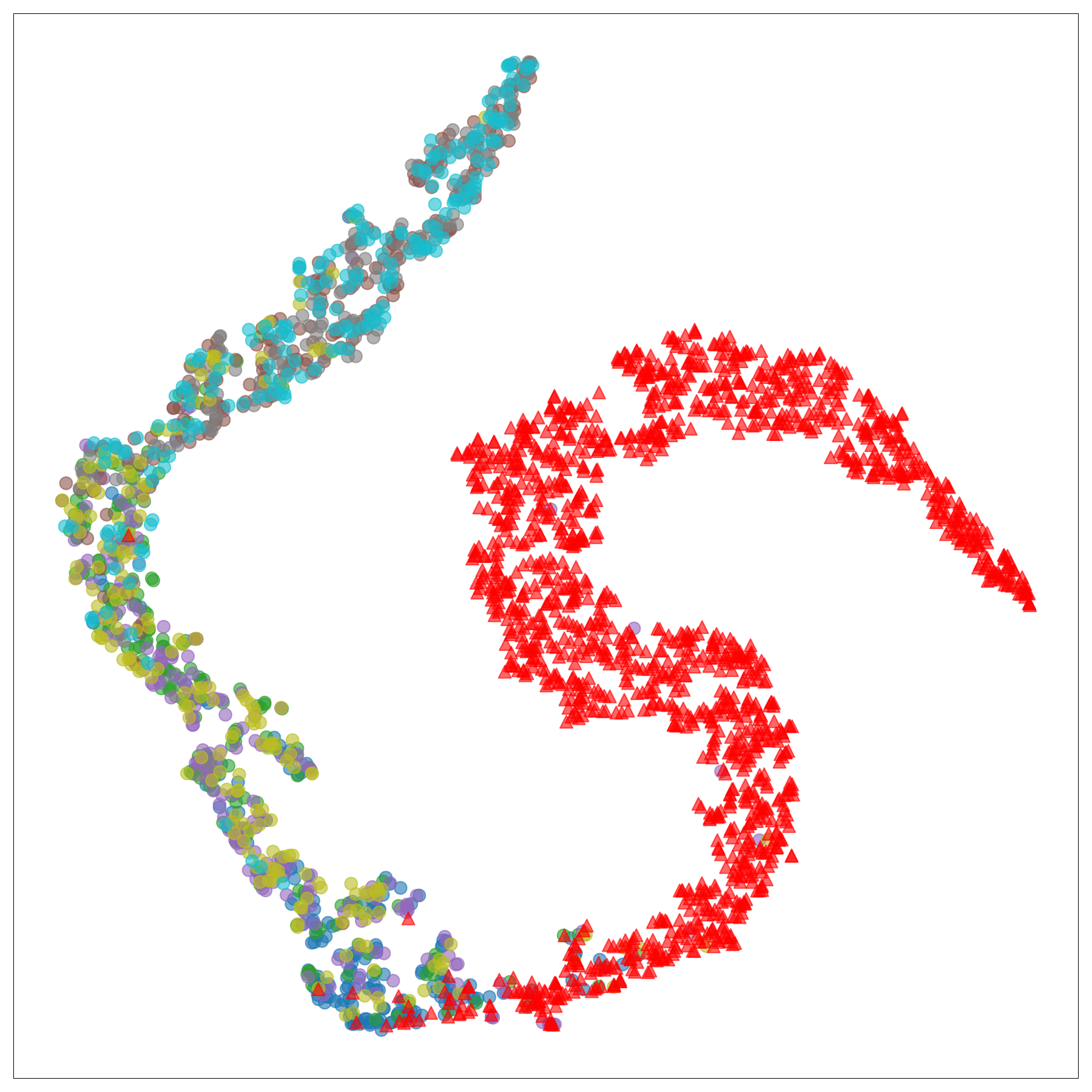}
  \caption{}
  \end{subfigure}
  
  \caption{(a) Pixel-level mapping GAN model t-SNE results. (b) NPR GAN model t-SNE results. (c) Pixel-level mapping Diffusion model t-SNE results. (d) NPR Diffusion model t-SNE results.}
  \label{fig:t-sne}
\end{figure}

\paragraph{Visualization of t-SNE Results.}
We present the t-SNE results of the fixed pixel-level mapping method trained using the aforementioned settings (ProGAN with 4 classes and SDv1.4) on both the GAN-generated and Diffusion-generated image sets in Fig. \ref{fig:t-sne}. We also reproduce the state-of-the-art method NPR using the same training and test setting. (a) and (b) utilize the Self-Synthesis dataset, which includes all GAN models, whereas (c) and (d) employ the GenImage dataset, comprising seven Diffusion models. The red triangles represent the features of real images, while the others correspond to the features of GAN-generated and Diffusion-generated images from the two datasets, respectively. The results demonstrate that the features extracted by mapping method effectively separate generated images from real images for both GAN and Diffusion models. This indicates the superior generalization capability of our approach.

\begin{figure}[t]
  \centering
  \includegraphics[width=1\linewidth]{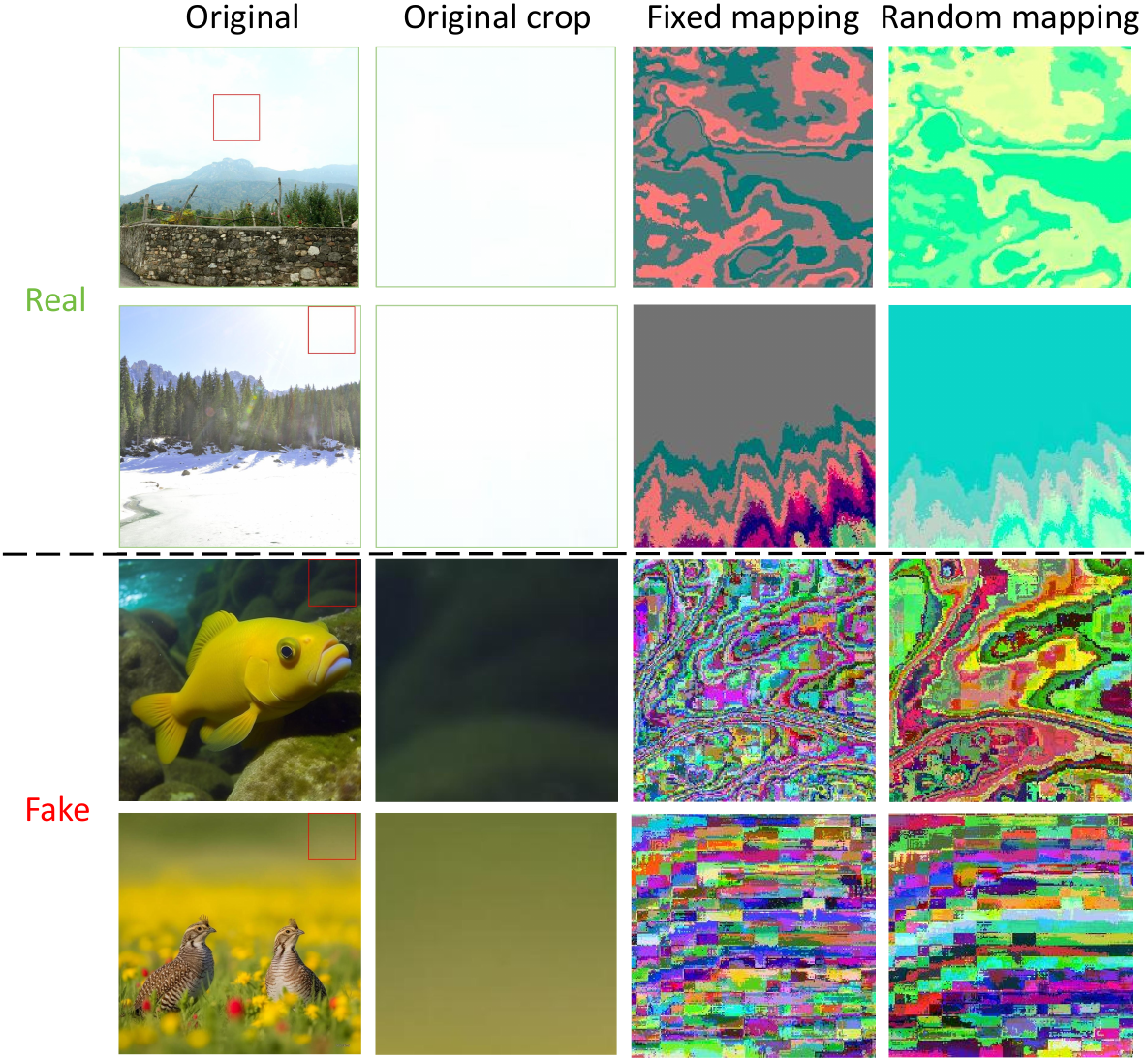}
  \caption{Visualization results of anomalies in mapped images.}
  \label{fig:amplify}
\end{figure}

\paragraph{Analysis of Anomalies in Mapped Images.}
Our pixel-level mapping analysis reveals distinctive artifacts in generated images, as demonstrated in Fig. \ref{fig:amplify}. For clearer visualization, we display results from the low-frequency components of high-resolution images. Using high-resolution datasets (Midjourney and RAISE \cite{dang2015raise}), we randomly crop low-frequency smooth regions for examination. The pixel-level mapping exposes abnormal checkerboard noise patterns which may originate from the upsampling process, while natural images maintain smooth transitions. Compared to original images, the mapped versions exhibit significantly reduced semantic features while amplifying anomalous traces, validating our method's capability to isolate generation-specific artifacts.

\begin{figure}[t]
  \centering
  \begin{subfigure}[t]{0.15\textwidth}
    \centering
  \includegraphics[width=\linewidth]{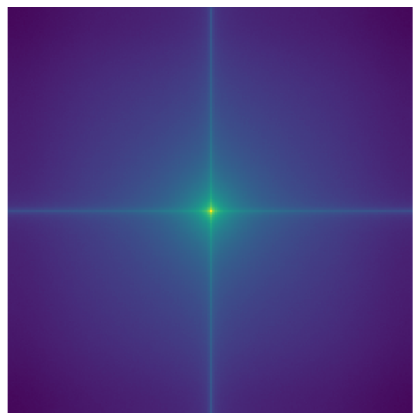}
  \caption{Original}
  \end{subfigure}
  \hfill
  \begin{subfigure}[t]{0.15\textwidth}
    \centering
  \includegraphics[width=\linewidth]{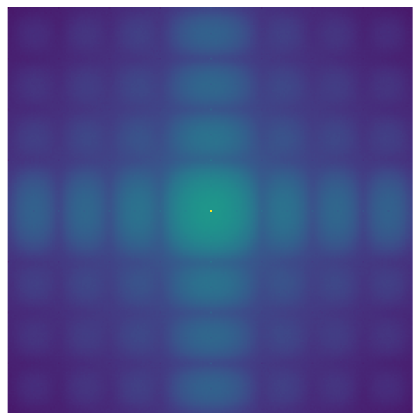}
  \caption{Shuffle (size=8)}
  \end{subfigure}
  \hfill
  \begin{subfigure}[t]{0.15\textwidth}
    \centering
  \includegraphics[width=\linewidth]{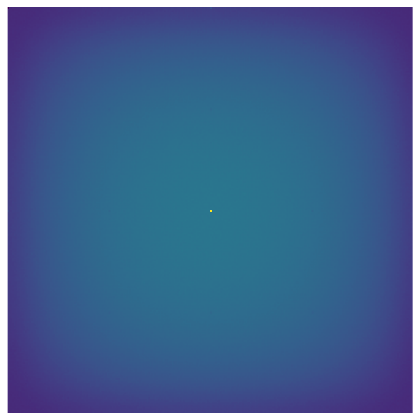}
  \caption{Shuffle (size=2)}
  \end{subfigure}
  \hfill
   \begin{subfigure}[t]{0.15\textwidth}
    \centering
  \includegraphics[width=\linewidth]{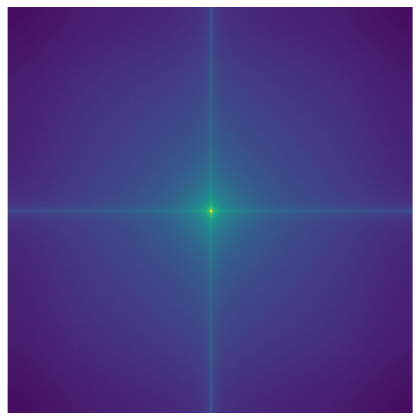}
  \caption{NPR}
  \end{subfigure}
  \hfill
   \begin{subfigure}[t]{0.15\textwidth}
    \centering
  \includegraphics[width=\linewidth]{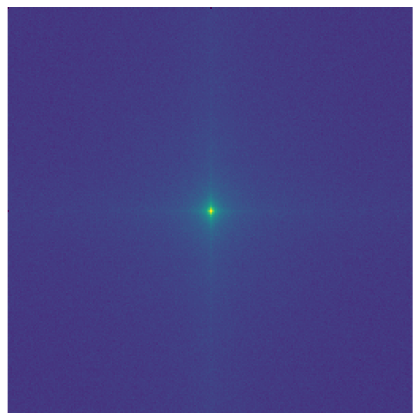}
  \caption{Fixed mapping}
  \end{subfigure}
  \hfill
   \begin{subfigure}[t]{0.15\textwidth}
    \centering
  \includegraphics[width=\linewidth]{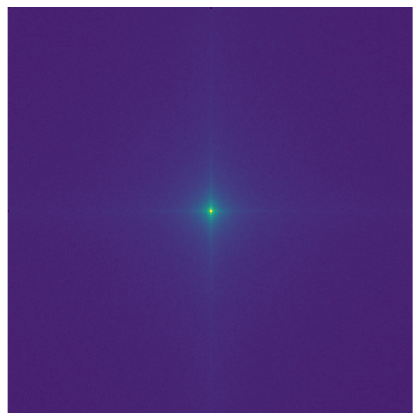}
  \caption{Random mapping}
  \end{subfigure}
  
  \caption{Frequency spectra of different semantic-reduction methods.}
  \label{fig:fre_comparison}
\end{figure}

\begin{figure}[t]
  \centering
  \includegraphics[width=0.8\linewidth]{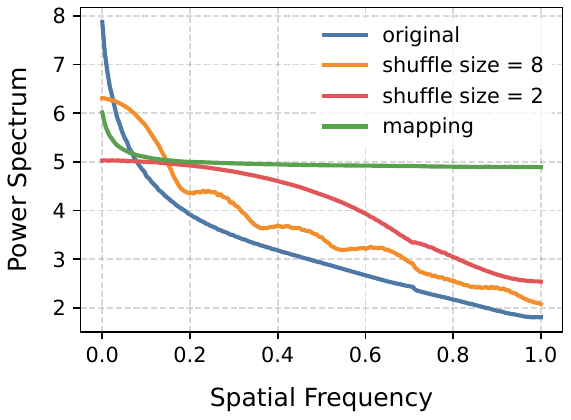}
  \caption{Power spectral representation of images under different processing methods.}
  \label{fig:spectral}
\end{figure}

\paragraph{Frequency Spectra Comparisons.}
We compare the spectral performance of different semantic-reduction methods by averaging 1,000 spectral images as shown in Figure \ref{fig:fre_comparison}, while the shuffle method narrows the low/high-frequency gap (particularly with patch size=2), it still exhibits clear energy drops at spectral boundaries. The NPR approach fails to effectively reduce the frequency gap through residual operation, whereas our mapping method significantly equalizes the energy distribution, enhancing high-frequency features for classifier training. We convert the 2D spectral data into 1D azimuthal integral spectrum \cite{durall2020watch}, which illustrates the spectral energy distribution across frequencies, with the horizontal axis ranging from low to high frequencies and the vertical axis representing spectral energy, as shown in Figure \ref{fig:spectral}. It reveals that although shuffling attenuates low-frequency power, it fails to enhance high-frequency components. Instead, the proposed method effectively narrows the spectral power gap between low- and high-frequency components and forces classifiers to prioritize high-frequency artifacts during training.

\section{Conclusion}
This paper presents a pixel-level mapping method that addresses classifier reliance on semantic bias by suppressing low-frequency features while enhancing high-frequency artifacts, significantly improving cross-model and cross-distribution detection generalization. The computationally efficient preprocessing step applies pixel-value transformations before classification to amplify inter-pixel differences and disrupt low-frequency biases. Extensive experiments on multiple benchmarks demonstrate the remarkable performance of the proposed method in generalization scenarios, offering a new direction for semantic bias reduction in synthetic image detection.

\section{Acknowledgments}
This work was supported by the Innovation Funding of Institute of Computing Technology, Chinese Academy of Sciences under Grant No. E561090 and E561160.

\bibliography{aaai2026}


\section{Appendix}
The appendix is organized as follows:
\begin{itemize}[leftmargin=*]
    \item Section A gives details in experimental settings and compares with SOTA methods on additional dataset.
    
    \item Section B shows experiments on post-processing shortcut analysis.
    
    \item Section C demonstrates more visualization comparisons of images processed by different semantic-reduction methods.
    
    \item Section D supplies additional comparative results on abnormal distribution in low-frequency component mappings.
\end{itemize}

\section{A. Experimental Details and Additional Comparison}
\begin{table*}[t]
  \centering
  \centering
   \resizebox{\textwidth}{!}{%
  \begin{tabular}{lcccccccccccccccc|cc}
    \toprule
    \multirow{2}{*}{Method} & 
    \multicolumn{2}{c}{Guided} & 
    \multicolumn{2}{c}{LDM\_200} & 
    \multicolumn{2}{c}{LDM\_200\_cfg} &
    \multicolumn{2}{c}{LDM\_100} &
    \multicolumn{2}{c}{GLIDE\_100\_27} &
    \multicolumn{2}{c}{GLIDE\_50\_27} &
    \multicolumn{2}{c}{GLIDE\_100\_10}&
    \multicolumn{2}{c}{DALLE}&
    \multicolumn{2}{c}{Mean} \\
    \cmidrule(lr){2-19}
  &  Acc. &  A.P. &  Acc. &  A.P. &  Acc. &  A.P. &  Acc. &  A.P. &  Acc. &  A.P. &  Acc. &  A.P. &  Acc. &  A.P. &  Acc. &  A.P. &  Acc. &  A.P. \\
    \midrule
CNN-Spot \cite{wang2020cnn} & 60.1 & 73.7 & 54.0 & 70.6 & 55.0 & 71.0 & 54.1 & 70.5 & 60.8 & 80.7 & 63.8 & 84.9 & 65.7 & 82.1 & 55.6 & 70.6 & 58.6 & 75.5\\
Patchfor \cite{chai2020makes} & 67.4 & 75 & 76.5 & 87.1 & 76.1 & 86.7 & 75.8 & 86.4 & 74.8 & 85.4 & 73.3 & 83.7 & 68.5 & 78.4 & 67.9 & 75.7 & 72.5 & 82.3\\
Co-occurence \cite{nataraj2019detecting} & 60.5 & 70.2 & 70.7 & 91.2 & 70.6 & 89 & 71.0 & 92.4 & 70.6 & 89.3 & 69.6 & 88.3 & 69.9 & 82.8 & 67.6 & 81 & 68.8 & 85.5\\
Freq-spec \cite{zhang2019detecting} & 50.9 & 57.7 & 50.4 & 77.7 & 50.4 & 77.2 & 50.3 & 76.5 & 51.7 & 68.6 & 51.4 & 64.6 & 50.4 & 61.9 & 50.0 & 67.8 & 50.7 & 69.0\\
F3Net \cite{qian2020thinking} & 69.2 & 70.5 & 68.2 & 73.8 & 75.4 & 81.7 & 68.8 & 74.6 & 81.7 & 89.8 & 83.2 & 91 & 83.1 & 90.9 & 66.3 & 71.8 & 74.5 & 80.5\\
UnivFD \cite{ojha2023towards} & 70.0 & 79.2 & 94.2 & 95.8 & 73.8 & 79.8 & 94.4 & 95.9 & 79.1 & 93.9 & 79.9 & 95.1 & 78.1 & 94.6 & 86.8 & 88.5 & 82.0 & 90.4\\
LGrad  \cite{tan2023learning} & 77.5 & 87.1 & 94.2 & 99 & 95.9 & 99.2 & 94.8 & 99.2 & 87.4 & 93.2 & 90.7 & 95.1 & 89.6 & 94.9 & 88.4 & 97.2 & 89.8 & 95.6\\
FreqNet \cite{tan2024frequency} & 86.7 & 96.3 & 84.6 & 96.1 & \textbf{99.6} & \textbf{100.0} & 65.6 & 62.3 & 85.7 & \underline{99.8} & \underline{97.4} & \underline{99.8} & 88.2 & 96.4 & 59.1 & 77.8 & 83.4 & 91.1\\
NPR \cite{tan2024rethinking} & 84.6 & 98.3 & 97.7 & \underline{99.9} & \underline{98.0} & \underline{99.9} & 98.2 & \underline{99.9} & 96.3 & \textbf{99.9} & 97.2 & \textbf{99.9} & \underline{97.4} & \textbf{99.9} & 87.2 & 99.3 & 94.6 & \underline{99.6}\\
FatFormer \cite{liu2024forgery} & 76.0 & 92 & \underline{98.6} & 99.8 & 94.9 & 99.1 & \underline{98.7} & \underline{99.9} & 94.4 & 99.1 & 94.7 & 99.4 & 94.2 & 99.2 & \underline{98.8} & \underline{99.8} & 93.8 & 98.5\\
C2P-CLIP \cite{tan2025c2p} & 69.1 & 94.1 & \textbf{99.3} & \textbf{100} & 97.3 & 99.8 & \textbf{99.3} & \textbf{100.0} & 95.3 & 99.7 & 95.3 & \underline{99.8} & 96.1 & \underline{99.8} & 98.6 & \textbf{99.9} & 93.8 & 99.2\\
\midrule
Fixed-mapping & \textbf{96.0} & \textbf{99.6} & 98.4 & 99.7 & \underline{98.0} & 99.7 & 98.1 & 99.7 & \textbf{98.5} & \textbf{99.9} & \textbf{99.5} & \textbf{99.9} & \textbf{99.6} & \textbf{99.9} & \textbf{99.2} & \underline{99.8} & \textbf{98.4} & \textbf{99.8}\\
Random-mapping & \underline{92.5} & \underline{98.7} & 97.8 & 99.4 & \underline{98.0} & 99.4 & 98.0 & 99.3 & \underline{97.4} & 99.4 & \underline{97.4} & 99.6 & 97.1 & 99.4 & 98.2 & 99.7 & \underline{97.0} & 99.4\\
\bottomrule
  \end{tabular}%
  }
  \caption{Cross-model detection accuracy(Acc.) and Average Precision(A.P.) results on the \textbf{UniversalFakeDetect}, where all models are trained exclusively on ForenSynths (ProGAN) data using the 4-class setting. \textbf{Bold} and \underline{underline} indicate top and runner-up performance respectively.}
  \label{universal-fake-detect-table-4class}
\end{table*}

\subsection{Experimental Details}
We provide detailed descriptions of the two test sets used in the main paper, as well as the additional test set included in the appendix.
\begin{itemize}[leftmargin=*]
\item \textbf{Self-Synthesis GAN dataset \cite{tan2024frequency}.}
The Self-Synthesis benchmark constitutes a specialized GAN dataset, incorporating nine state-of-the-art architectures (AttGAN \cite{he2019attgan}, BEGAN \cite{berthelot2017began}, CramerGAN \cite{bellemare2017cramer}, InfoMaxGAN \cite{lee2021infomax}, MMDGAN \cite{li2017mmd}, RelGAN \cite{nie2018relgan}, S3GAN \cite{luvcic2019high}, SNGAN \cite{miyato2018spectral}, and STGAN \cite{liu2019stgan}) to extend evaluation scenarios. We adopt all the 9 GAN models as test set. This curated collection establishes a rigorous benchmark for assessing generalization capabilities of AI-generated image detectors.

\item \textbf{GenImage dataset \cite{zhu2023genimage}.}
The GenImage benchmark incorporates synthetic images from eight advanced generative architectures, comprising seven diffusion models (Midjourney, SDv1.4 \cite{rombach2022high}, SDv1.5 \cite{rombach2022high}, ADM \cite{dhariwal2021diffusion}, GLIDE \cite{nichol2021glide}, Wukong, VQDM \cite{egiazarian2024accurate}) and BigGAN \cite{brock2018large}. We adopt all the 8 models as test set. It surpasses conventional benchmarks in both visual fidelity and detection complexity, presenting heightened challenges for existing forensic methods. 

\item \textbf{UniversalFakeDetect dataset \cite{ojha2023towards}.} For this benchmark, we exclusively utilize four diffusion models (Guided 
 \cite{dhariwal2021diffusion}, LDM \cite{rombach2022high}, GLIDE \cite{nichol2021glide}, and DALL-E \cite{ramesh2022hierarchical}) collected by the authors as the test set. Distinct sampling strategies are implemented for specific DMs (specifically LDM \cite{rombach2022high} and GLIDE \cite{nichol2021glide}). Multi-source real images from LAION \cite{schuhmann2021laion} and ImageNet datasets \cite{russakovsky2015imagenet} are integrated to systematically assess detector performance.
\end{itemize}

\paragraph{Reported results on GenImage dataset.}
The results of accuracy shown in GenImage dataset are reported from different works. The results of ResNet-50, DeiT-S, Swin-T, CNNSpot, Spec, F3Net, and GramNet are from GenImage \cite{zhu2023genimage}. The results of UnivFD, FreqNet, NPR, FatFormer, C2P-CLIP are from C2P-CLIP \cite{tan2025c2p}. The results of DIRE and DRCT are from DRCT \cite{chen2024drct}.

\subsection{Additional Comparison}
\paragraph{Evaluation on UniversalFakeDetect dateset.}
The results of accuracy are shown in Table \ref{universal-fake-detect-table-4class}. The results of relative methods are from C2P-CLIP \cite{tan2025c2p}. All the detection models are trained on the training set of ForenSynths (4 classes) \cite{wang2020cnn} with a single generative model ProGAN which is mentioned in the source of results \cite{tan2025c2p}. The proposed pixel-level mapping method achieves a new state-of-the-art result with an average accuracy of 98.4\%. Our fixed pixel-level mapping approach surpasses the baseline UniFD \cite{ojha2023towards} by 16.4\% in classification accuracy and outperforms the current state-of-the-art NPR \cite{tan2024rethinking} by 3.8\%, demonstrating that reducing semantic bias while shifting spectral energy from low-frequency to high-frequency components enables convolutional networks to better focus on high-frequency forgery traces. The random pixel-level mapping variant also achieves significant improvements, exceeding UniFD and NPR baselines by 15\% and 2.4\% respectively, indicating that employing arbitrary mapping strategies effectively enhances the model's capability to extract generalized forgery artifacts.

\begin{table*}[t]
  \resizebox{\textwidth}{!}{%
  \begin{tabular}{lcccccccccc}
    \toprule
    \multirow{3}{*}{Method} & \multirow{3}{*}{Venue} & \multirow{3}{*}{Guided} & \multicolumn{3}{c}{LDM} & \multicolumn{3}{c}{GLIDE} &\multirow{3}{*}{DALLE}&\multirow{3}{*}{mAcc} \\
\cmidrule(lr){4-6}\cmidrule(lr){7-9}
 &  &  & 200 & 200 & 100 & 100 & 50 & 100 \\
 &  &  & steps & w/cfg &  steps & 27 & 27 & 10 \\
    \midrule
    CNN-Spot \cite{wang2020cnn}     & CVPR2020 & 60.1 & 54.0 & 55.0 & 54.1 & 60.8 & 63.8 & 65.7 & 55.6 & 58.6\\
    Patchfor \cite{chai2020makes}     & ECCV2020 & 67.4 & 76.5 & 76.1 & 75.8 & 74.8 & 73.3 & 68.5 & 67.9 & 72.5\\
    Co-occurence \cite{nataraj2019detecting}& Elect. Imag.   & 60.5 & 70.7 & 70.6 & 71.0 & 70.6 & 69.6 & 69.9 & 67.6 & 68.8\\
    Freq-spec \cite{zhang2019detecting}   & WIFS2019 & 50.9 & 50.4 & 50.4 & 50.3 & 51.7 & 51.4 & 50.4 & 50.0 & 50.7\\
    F3Net \cite{qian2020thinking}  & ECCV2020 & 69.2 & 68.2 & 75.4 & 68.8 & 81.7 & 83.25 & 83.1 & 66.3 & 74.5\\
    UnivFD \cite{ojha2023towards} & CVPR2023 & 70.0 & 94.2 & 73.8 & 94.4 & 79.1 & 79.9 & 78.1 & 86.8 & 82.0\\
    LGrad  \cite{tan2023learning}  & CVPR2023 & 77.5 & 94.2 & 95.9 & 94.8 & 87.4 & 90.7 & 89.6 & 88.4 & 89.8\\
    FreqNet \cite{tan2024frequency}  & AAAI2024 & 86.7 & 84.6 & 99.6 & 65.6 & 85.7 & 97.4 & 88.2 & 59.1 & 83.4\\
    NPR \cite{tan2024rethinking}      & CVPR2024 & 84.6 & 97.7 & 98.0 & 98.2 & 96.3 & 97.2 & 97.4 & 87.2 & 94.6\\
    FatFormer \cite{liu2024forgery} & CVPR2024 & 76.0 & 98.6 & 94.9 & 98.7 & 94.4 & 94.7 & 94.2 & 98.8 & 93.8\\
    C2P-CLIP \cite{tan2025c2p} & AAAI2025 & 69.1 & 99.3 & 97.3 & 99.3 & 95.3 & 95.3 & 96.1 & 98.6 & 93.8\\
    \midrule
    Fixed-mapping& Ours & 97.8 & 97.1 & 98.0 & 98.5 & 98.3 & 98.0 & 89.1 & 76.2 & 94.1\\
    Random-mapping& Ours & 98.5 & 97.8 & 98.1 & 98.4 & 98.6 & 98.2 & 85.9 & 80.0 & 94.4\\

    \bottomrule
  \end{tabular}%
  }
  \caption{Cross-model detection accuracy (\%) results on the \textbf{UniversalFakeDetect}. }
  \label{tab:bias1}
  \centering
\end{table*}

\begin{table*}[t]
\resizebox{\textwidth}{!}{%
\begin{tabular}{lcccccccccc}
\toprule
Method & AttGAN & BEGAN & CramerGAN & InfoMaxGAN & MMDGAN & RelGAN & S3GAN & SNGAN & STGAN & mAcc \\
\midrule
CNNDetection \cite{wang2020cnn} & 51.1 & 50.2 & 81.5 & 71.1 &  72.9 & 53.3 & 55.2 &  62.7 &  63.0 & 62.3\\
Frank \cite{frank2020leveraging} & 65.0 & 39.4 & 31.0 & 41.1 & 38.4 & 69.2 & 69.7 & 48.4 & 25.4 & 47.5 \\
Durall \cite{durall2020watch} & 39.9 & 48.2 & 60.9 & 50.1 & 59.5 & 80.0 & 87.3 & 54.8 & 62.1 & 60.3 \\
Patchfor \cite{chai2020makes} & 68.0 &  97.1 &  97.8 &  93.6 & 97.9 & 99.6 &  66.8 &   97.6 & 92.7 &  90.1\\
F3Net \cite{qian2020thinking} & 85.2 & 87.1 & 89.5 & 67.1 & 73.7 & 98.8 & 65.4 & 51.6  & 60.3 & 75.4  \\
SelfBlend \cite{shiohara2022detecting} &  63.1 & 56.4 & 75.1 & 79.0 & 68.6 & 73.6 & 53.2  &61.6 & 61.2 & 65.8 \\
GANDetection \cite{mandelli2022detecting} &  57.4&  67.9  &67.8 &67.6 & 67.7& 60.9 &69.6& 66.7&  69.6 & 66.1 \\
LGrad \cite{tan2023learning}  &  68.6 & 69.9 & 50.3 & 71.1& 57.5 &89.1 & 78.5& 78.0 & 54.8 & 68.6  \\
UnivFD \cite{ojha2023towards} & 78.5 & 72.0 & 77.6 & 77.6 & 77.6  &78.2 & 85.2 & 77.6 & 74.2 & 77.6  \\
NPR \cite{tan2024rethinking} &  83.0 & 99.0 & 98.7 & 94.5& 98.6& 99.6& 79.0& 88.8& 98.0& 93.2 \\
\midrule
Fixed-mapping&99.8 & 90.2 & 97.8 & 94.0 & 98.4 & 99.6 & 77.3 & 97.9 & 99.6 &  95.0 \\
Random-mapping& 99.8 & 99.0 & 96.8 & 89.6 & 97.5 & 99.7& 73.8 & 97.0 & 99.8 & 94.8\\
\bottomrule
\end{tabular}%
  }
\caption{Cross-GAN performance analysis using the \textbf{Self-Synthesis} 9 GANs dataset.}
\label{tab:bias2}
\centering
\end{table*}

\begin{table*}[t]
\resizebox{\textwidth}{!}{%
\begin{tabular}{lcccccccccc}
\toprule
Method & Ref & Midjourney & SDv1.4 & SDv1.5 & ADM & GLIDE & Wukong & VQDM & BigGAN & mAcc \\
\midrule
ResNet-50 \cite{he2016deep} & CVPR2016 & 54.9 & 99.9 & 99.7 & 53.5 & 61.9 & 98.2 & 56.6 & 52.0 & 72.1 \\
DeiT-S \cite{touvron2021training} & ICML2021 & 55.6 & 99.9 & 99.8 & 49.8 & 58.1 & 98.9 & 56.9 & 53.5 & 71.6 \\
Swin-T \cite{liu2021swin} & ICCV2021 & 62.1 & 99.9 & 99.8 & 49.8 & 67.6 & 99.1 & 62.3 & 57.6 & 74.8 \\
CNNSpot \cite{wang2020cnn} & CVPR2020 & 52.8 & 96.3 & 95.9 & 50.1 & 39.8 & 78.6 & 53.4 & 46.8 & 64.2 \\
Spec \cite{zhang2019detecting} & WIFS2019 & 52.0 & 99.4 & 99.2 & 49.7 & 49.8 & 94.8 & 55.6 & 49.8 & 68.8 \\
F3Net \cite{qian2020thinking} & ECCV2020 & 50.1 & 99.9 & 99.9 & 49.9 & 50.0 & 99.9 & 49.9 & 49.9 & 68.7 \\
GramNet \cite{liu2020global} & CVPR2020 & 54.2 & 99.2 & 99.1 & 50.3 & 54.6 & 98.9 & 50.8 & 51.7 & 69.9 \\
UnivFD \cite{ojha2023towards} & CVPR2023 & 93.9 & 96.4 & 96.2 & 71.9 & 85.4 & 94.3 & 81.6 & 90.5 & 88.8 \\
DIRE \cite{wang2023dire} & ICCV2023 & 50.4 & 100.0 & 99.9 & 52.3 & 67.2 & 100.0 & 50.1 & 50.0 & 71.2 \\
FreqNet \cite{tan2024frequency} & AAAI2024 & 89.6 & 98.8 & 98.6 & 66.8 & 86.5 & 97.3 & 75.8 & 81.4 & 86.8 \\
NPR \cite{tan2024rethinking} & CVPR2024 & 81.0 & 98.2 & 97.9 & 76.9 & 89.8 & 96.9 & 84.1 & 84.2 & 88.6 \\
FatFormer \cite{liu2024forgery} & CVPR2024 & 92.7 & 100.0 & 99.9 & 75.9 & 88.0 & 99.9 & 98.8 & 55.8 & 88.9 \\
DRCT \cite{chen2024drct} & ICML2024 & 91.5 & 95.0 & 94.4 & 79.4 & 89.2 & 94.7 & 90.0 & 81.7 & 89.5 \\
C2P-CLIP \cite{tan2025c2p} & AAAI2025 & 88.2 & 90.9 & 97.9 & 96.4 & 99.0 & 98.8 & 96.5 & 98.7 & 95.8 \\
VIB-Net \cite{zhang2025towards} & CVPR2025 & 88.1 & 99.6 & 99.2 & 73.9 & 74.3 & 98.3 & 89.4 & 91.2 & 89.3 \\
B-Free \cite{guillaro2025bias} & CVPR2025 & 89.0 & 93.5 & 93.5 & 92.7 & 83.3 & 94.1 & 90.8 & 94.6 & 91.4 \\
Effort \cite{yan2024effort} & ICML2025 & 82.4 & 99.8 & 99.8 & 78.7 & 93.3 & 97.4 & 91.7 & 77.6 & 91.1 \\
\midrule
Fixed-mapping & Ours& 88.2 & 91.9 & 92.1 & 90.7 & 89.2 & 91.0 & 92.2 & 88.9 &  90.5\\
Random-mapping & Ours & 91.5 & 89.5 & 92.3 & 88.2 & 91.4 & 90.5 & 93.3 & 90.1 & 90.9\\
\bottomrule
\end{tabular}%
  }
\caption{Accuracy (Acc) comparison across models tested on \textbf{GenImage}.}
\label{tab:bias3}
\centering
\end{table*}

\section{B. Discussion on Common Post-processing Biases}
Some findings \cite{grommelt2024fake, ricker2024aeroblade} have identified that common post-processing operations (e.g., JPEG compression, image resizing) leave distinctive traces in data distributions. Since most publicly available datasets are collected from internet sources, such post-processing artifacts are prevalent - for instance, generated images typically use PNG format while real images often exhibit JPEG compression artifacts. Notably, some detection methods inadvertently establish erroneous detection paradigms by relying on these post-processing traces rather than generation artifacts.

To verify that our proposed method learns genuine generation artifacts rather than spurious post-processing cues (which predominantly reside in high-frequency domains), we conduct controlled experiments using the RAISE \cite{dang2015raise} dataset as our pristine real image source. RAISE contains high-resolution, camera-original images without any post-processing. We train an ADM \cite{dhariwal2021diffusion} model exclusively on randomly cropped RAISE patches, following Corvi et al.'s finding \cite{corvi2023intriguing} that generative models replicate all statistical distributions present in their training data. Since our ADM training data contains no post-processing artifacts, the generated images should only contain model-specific generation traces. We trained our classifier on ADM-generated images versus RAISE real images, while keeping the experimental settings unchanged with those of other methods.

As shown in Tab. \ref{tab:bias1}, Tab. \ref{tab:bias2} and Tab. \ref{tab:bias3}, the proposed method demonstrates consistent performance across three public benchmarks. The high resolution of RAISE dataset leads to substantial semantic discrepancy between randomly cropped patches and the test set, resulting in decreased classifier accuracy on lower-resolution tests. Yet, this confirms that our approach detects genuine generation traces instead of post-processing shortcuts, while maintaining generalizability across both GAN and diffusion models. This provides empirical validation for our approach's effectiveness in learning authentic generation artifacts.

\section{C. Visualizations of Semantic-reduction Methods}
\begin{figure*}[t]
  \centering
  \includegraphics[width=1\linewidth]{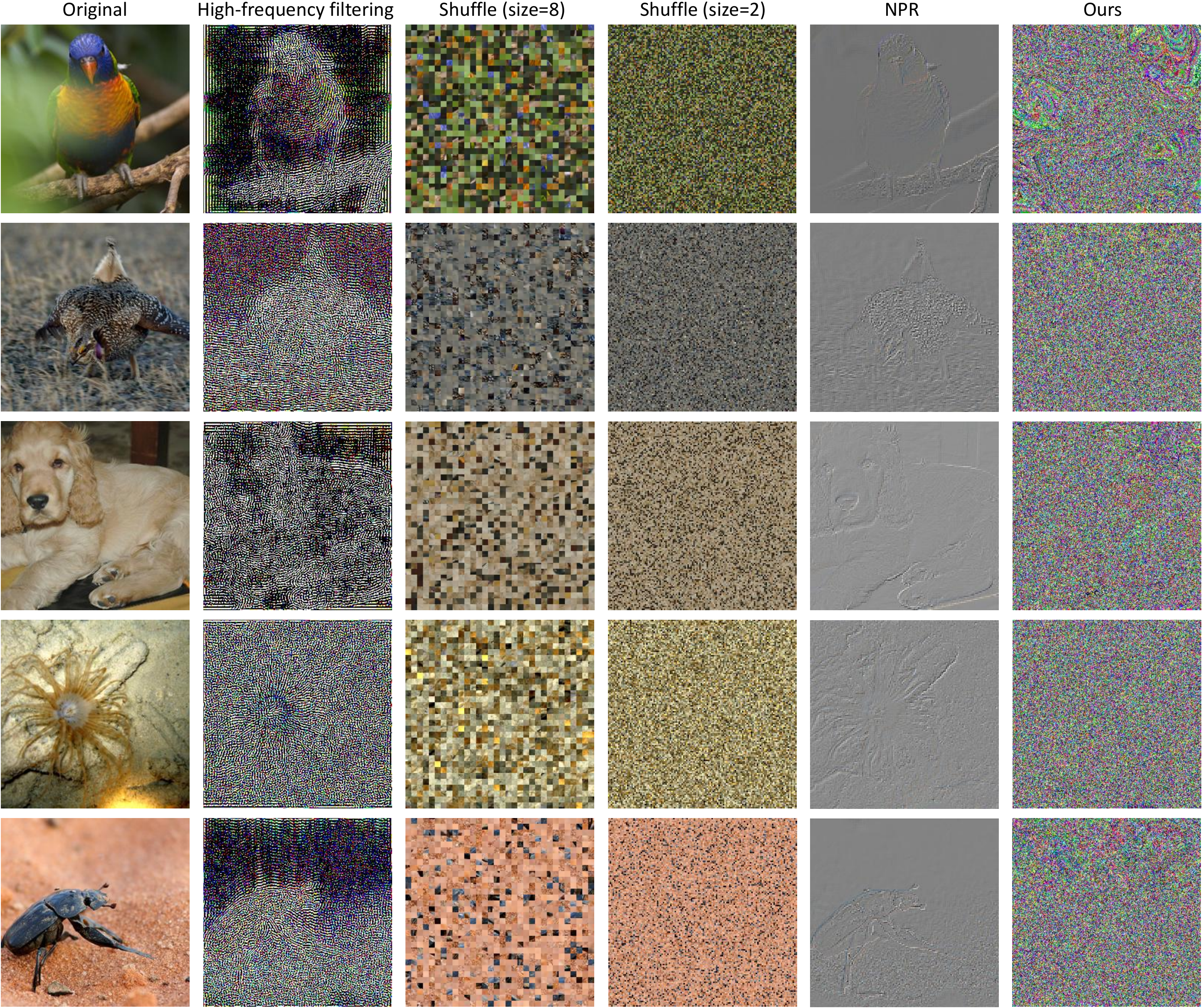}
  \caption{Additional visualization results of various semantic-reduction methods.}
  \label{fig:appendix_reduction}
\end{figure*}

As shown in Figure \ref{fig:appendix_reduction}, we compare the semantic-reduction effects of high-frequency filtering, patch shuffling, NPR, and our proposed pixel mapping method. The results demonstrate that while high-frequency filtering disrupts low-frequency components, significant semantic information remains discernible from high-frequency features like edges, indicating that only discarding low-frequency spectral can not reduce semantic information completely. The patch shuffling approach exhibits progressively stronger semantic disruption as the patch size decreases, though our ImageNet experiments reveal that classifiers can still learn semantic-related patterns from shuffled patch images. The NPR method extracts upsampling information through residual operations, yet retains noticeable semantic traces similar to filtering. In contrast, our method effectively transforms low-frequency semantic information into high-frequency components, substantially reducing the influence of low-frequency semantic bias.

\begin{figure*}[t]
  \centering
  \includegraphics[width=1\linewidth]{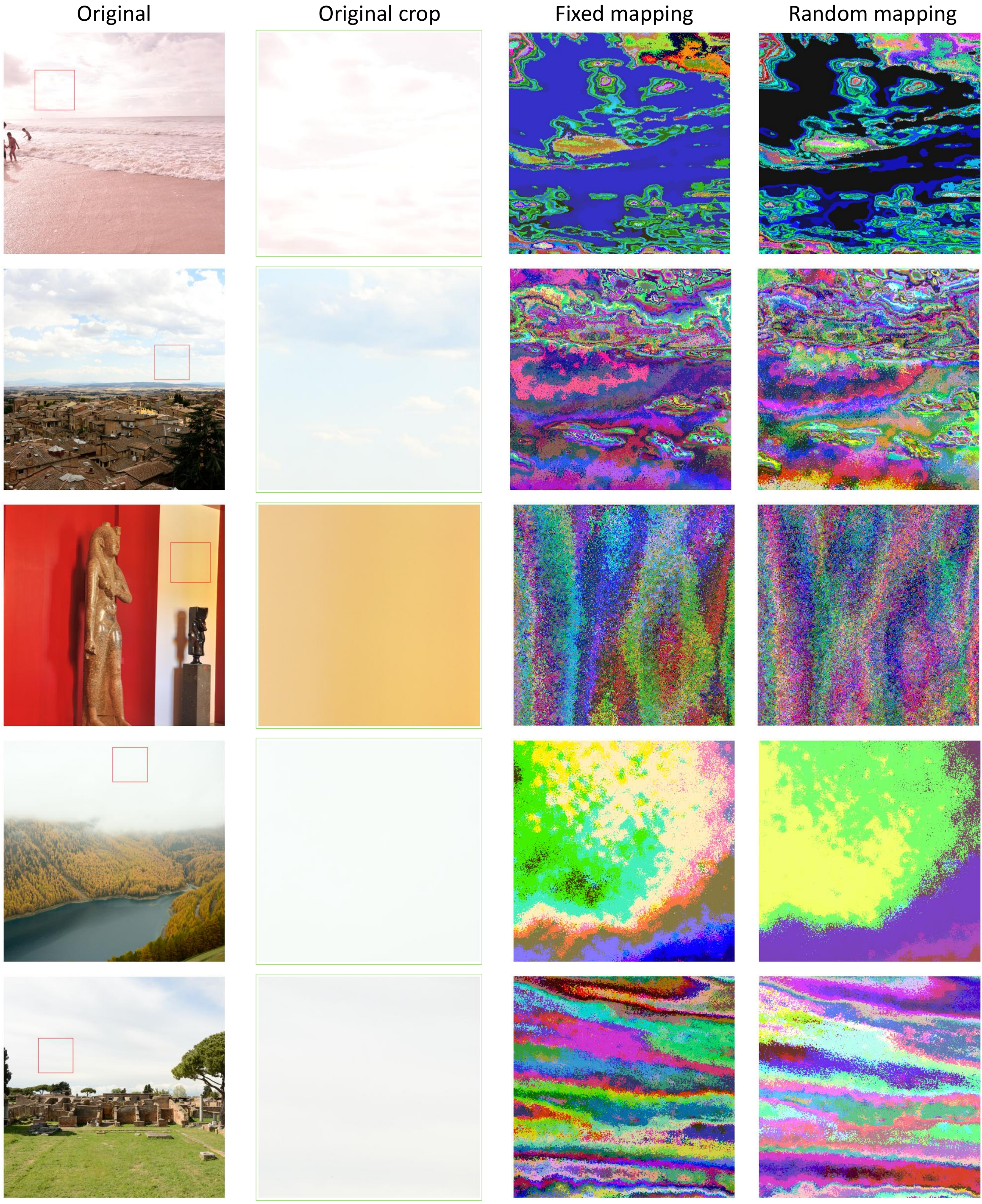}
  \caption{Anomaly distribution analysis of RAISE images through pixel-level mapping.}
  \label{fig:appendix_smooth_real}
\end{figure*}

\begin{figure*}[t]
  \centering
  \includegraphics[width=1\linewidth]{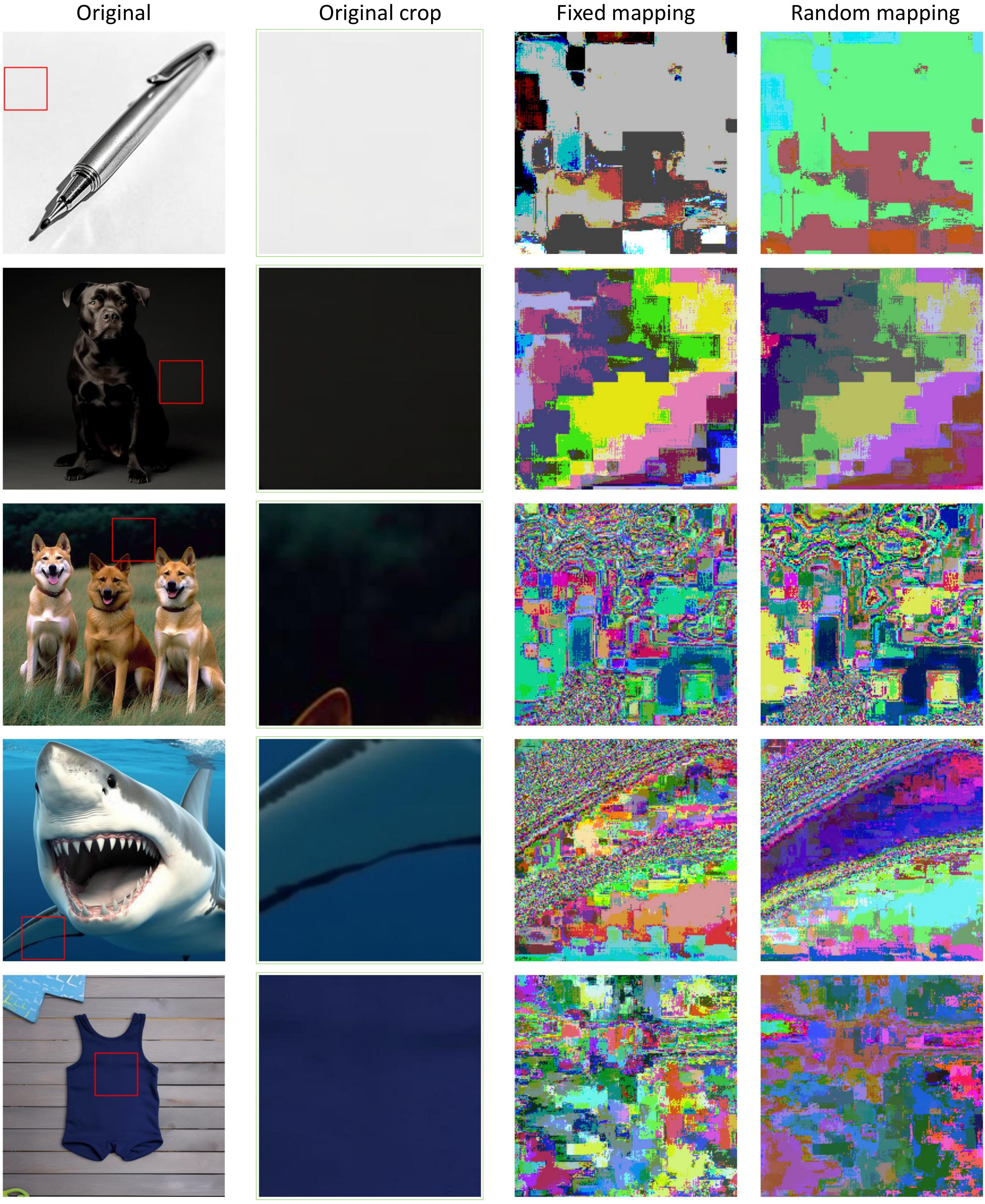}
  \caption{Anomaly distribution analysis of Midjourney images through pixel-level mapping.}
  \label{fig:appendix_smooth_fake}
\end{figure*}

\section{D. Visualizations of Abnormal Mapping Distribution}
Our proposed pixel mapping method disrupts the monotonic arrangement of adjacent pixels to convert low-frequency smooth regions into high-frequency components, thereby amplifying generation artifacts. To visualize this enhancement effect, Figures \ref{fig:appendix_smooth_real} and \ref{fig:appendix_smooth_fake} demonstrate the results after applying pixel mapping to low-frequency regions of real (RAISE dataset) and fake (Midjourney) images, respectively. For clearer visualization, we specifically selected high-resolution images for displaying. As shown in Figure \ref{fig:appendix_smooth_real}, real images maintain smooth stripe-like pixel distributions after mapping, whereas Figure \ref{fig:appendix_smooth_fake} reveals abnormal checkerboard patterns in fake images' low-frequency regions - likely caused by upsampling operations during image generation. These comparative results demonstrate that our pixel mapping method effectively amplifies generation traces in low-frequency components while reducing semantic bias in classifier training.

\end{document}